\documentclass[10pt,twocolumn,letterpaper]{article}
\usepackage[linesnumbered,ruled,vlined]{algorithm2e}

\usepackage{cvpr}
\usepackage{cvpr24}
\def\paperID{2001}
\def\confName{CVPR}
\def\confYear{2024}
\makeatletter
\renewcommand{\paragraph}{%
  \@startsection{paragraph}{4}%
  {\z@}{0ex \@plus 0ex \@minus 0ex}{-1em}%
  {\hskip\parindent\normalfont\normalsize\bfseries}%
}
\makeatother

\definecolor{myblue}{RGB}{151,198,194}
\definecolor{cvprblue}{rgb}{0.21,0.49,0.74}
\definecolor{textblue}{RGB}{151,198,194}
\definecolor{textgreen}{RGB}{181,209,177}
\definecolor{textred}{RGB}{147,44,80}
\definecolor{gb}{RGB}{68,179,110}
\definecolor{rb}{RGB}{147,44,80}
\definecolor{myred}{named}{Crimson}
\definecolor{myyellow}{named}{Goldenrod}
\newcommand{\gr}[1]{\textcolor{LimeGreen}{#1}}
\newcommand{\rr}[1]{\textcolor{textred}{#1}}

\def\model{\texttt{AnySkill}\xspace}
\acrodef{vlm}[VLM]{Vision-Language Model}
\acrodef{gail}[GAIL]{generative adversarial imitation learning}

\newcommand\numberthis{\addtocounter{equation}{1}\tag{\theequation}}

\newcommand{\Ra}[1]{\textcolor{myred}{\textbf{R1}}}
\newcommand{\Rb}[1]{\textcolor{myblue}{\textbf{R2}}}
\newcommand{\Rc}[1]{\textcolor{JungleGreen}{\textbf{R3}}}

\title{\model: Learning Open-Vocabulary Physical Skill for Interactive Agents\vspace{-12pt}}
\author{%
    Jieming Cui$^{1,2*}$, Tengyu Liu$^{2*}$, Nian Liu$^{2,3*}$, Yaodong Yang$^{1}$, Yixin Zhu$^{1,~\textrm{\Letter}}$, Siyuan Huang$^{2,~\textrm{\Letter}}$
    \vspace{6pt}\\
    \small $^1$ Institute for Artificial Intelligence, Peking University\quad{}
    $^2$ National Key Laboratory of General Artificial Intelligence, BIGAI\\
    \small $^3$ School of Artificial Intelligence, Beijing University of Posts and Telecommunications
    \vspace{6pt}\\
    \href{https://anyskill.github.io}{https://anyskill.github.io}
    \vspace{-12pt}
}
\begin{document}

\twocolumn[{
\renewcommand\twocolumn[1][]{#1}
\maketitle
\vspace{-12pt}
\begin{center}
    \captionsetup{type=figure}
    \includegraphics[width=\linewidth]{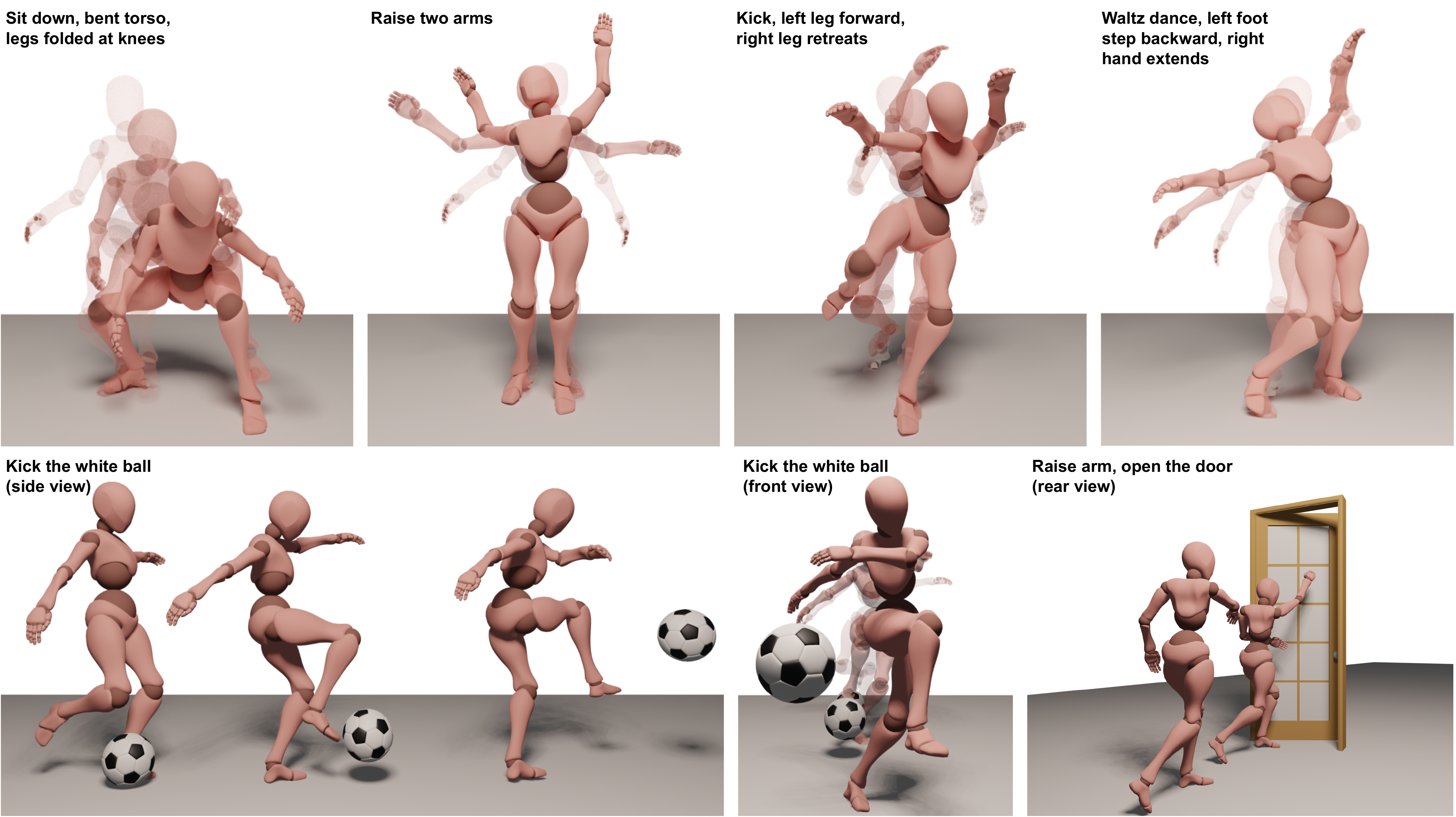}
    \captionof{figure}{\textbf{Diverse motions generated by \model{} conditioned on various instructions.} When provided with an open-vocabulary text description of a motion, \model{} is adept at learning natural and flexible motions that closely align with the description, facilitated by an image-based reward mechanism. Additionally, \model{} demonstrates proficiency in learning interactions with dynamic objects, showcasing its versatile motion generation capabilities.}
    \label{fig:teaser}
\end{center}
}]

\begin{abstract}
\vspace{-12pt}
Traditional approaches in physics-based motion generation, centered around imitation learning and reward shaping, often struggle to adapt to new scenarios. To tackle this limitation, we propose \model, a novel hierarchical method that \emph{learns physically plausible interactions following open-vocabulary instructions}. Our approach begins by developing a set of atomic actions via a low-level controller trained via imitation learning. Upon receiving an open-vocabulary textual instruction, \model employs a high-level policy that selects and integrates these atomic actions to maximize the CLIP similarity between the agent's rendered images and the text. An important feature of our method is the use of image-based rewards for the high-level policy, which allows the agent to \emph{learn interactions with objects without manual reward engineering}. We demonstrate \model's capability to generate realistic and natural motion sequences in response to unseen instructions of varying lengths, marking it the first method capable of open-vocabulary physical skill learning for interactive humanoid agents.
\vspace{-12pt}
\end{abstract}

\section{Introduction}

Confronted with a soccer ball, an individual might engage in various actions such as kicking, dribbling, passing, or shooting. This interaction capability is feasible even for someone who has only observed soccer games, never having played. This ability exemplifies the human aptitude for learning open-vocabulary physical interaction skills from visual experiences and applying these skills to novel objects and actions. Equipping interactive agents with this capability remains a significant challenge.

Recent physical skill learning methods predominantly rely on imitation learning to acquire realistic physical motions and interactions~\cite{peng2021amp,peng2018deepmimic}. However, this approach limits their adaptability to unforeseen scenarios with novel instructions and environments. Furthermore, neglecting physical laws in current models leads to unnatural and unrealistic motions, such as floating, penetration, and foot sliding, despite attempts to integrate physics-based penalties like gravity~\cite{xu2023language,zhang2023motiongpt} and collision~\cite{zhao2023synthesizing,xu2023interdiff,huang2023diffusion}. Enhancing the generalizability of physically constrained motion generation is essential for decreasing reliance on specific datasets and fostering a more profound comprehension of the world.

On top of generalizability, the ultimate goal is to generate natural and interactive motions from any text input, known as achieving open vocabulary, which significantly increases the complexity of the problem. Several studies have explored open-vocabulary motion generation using large-scale pretrained models~\cite{rocamonde2023vision,kumar2023words,hong2022avatarclip,tevet2022motionclip}. However, these models struggle to produce natural motions, particularly interactive motions that require understanding broader environmental contexts or object interactions~\cite{kumar2023words,hong2022avatarclip,tevet2022motionclip}. 

We identify a gap in motion generalizability on novel tasks and interaction capabilities with environments, hypothesizing that this is due to the reliance on improvised state representations and manually crafted reward mechanisms in prior works.
Inspired by the human ability to learn new physical skills from visual inputs, we propose utilizing a \ac{vlm} to offer flexible and generalizable state representations and image-based rewards for open-vocabulary skill learning.
We introduce \model{}, a hierarchical framework designed to equip virtual agents with the ability to learn open-vocabulary physical interaction skills. \model{} combines a shared low-level controller with a high-level policy tailored to each instruction, learning a repertoire of latent atomic actions through \ac{gail}, following CALM~\cite{tessler2023calm}. This ensures the naturalness and physical plausibility of each action. Then, for any open-vocabulary textual instruction, a high-level control policy dynamically selects latent atomic actions to optimize the CLIP~\cite{radford2021learning} similarity between the agent's rendered images and the textual instruction. This policy maintains physical plausibility and allows the agent to act according to a broad range of textual instructions. By leveraging CLIP similarity as a flexible and straightforward reward mechanism, our approach overcomes environmental limitations, facilitating interaction with any object. Despite the advances, creating natural and interactive actions for open-vocabulary models remains an ongoing challenge.

Extensive experiments demonstrate \model{}'s ability to execute physical and interactive skills learned from open-vocabulary instructions; \cref{fig:teaser} showcases various interactive and non-interactive examples. We further prove that our method outperforms existing open-vocabulary motion generation approaches in creating interaction motions.
 
To summarize, our contributions are three-fold:
\begin{itemize}[leftmargin=*,nolistsep,noitemsep]
    \item We introduce \model{}, a hierarchical approach that combines a low-level controller with a high-level policy, specifically designed for the learning of open-vocabulary physical skills.
    \item We leverage the \ac{vlm} (\ie, CLIP) to provide a novel means of generating flexible and generalizable image-based rewards. This approach eliminates the need for manually engineered rewards, facilitating the learning of both individual and interactive actions.
    \item Through extensive experimentation, we demonstrate that our method significantly surpasses existing approaches in both qualitative and quantitative measures. Importantly, \model{} empowers agents with the ability to engage in smooth and natural interactions with dynamic objects across a variety of contexts.
\end{itemize}

\section{Related Work}

\textbf{Physical skills learning} emphasizes mastering motions that adhere to physical laws, including gravity, friction, and penetration. This domain has seen approaches that either employ specific loss functions to address constraints like foot-ground penetration~\cite{yuan2023physdiff}, body-object interaction~\cite{chen2019holistic++,wang2021synthesizing,wang2021scene,yi2023mime,zhang2020generating,zhao2022compositional,lee2023locomotion,tseng2023edge,tripathi20233d,hassan2021stochastic,jiang2023full,qi2018human,cui2024probio,wang2024move}, self-collision~\cite{mihajlovic2022coap,tian2023recovering,khazoom2022humanoid}, and gravity~\cite{xie2021physics,gartner2022differentiable,salzmann2022motron}, or leverage physics simulators~\cite{todorov2012mujoco,makoviychuk2021isaac,peng2021amp,juravsky2022padl,tessler2023calm,peng2022ase} for more dynamic fidelity. Despite these efforts, ensuring fine-grained physical plausibility, especially in complex interactions, remains a challenge. The integration of reinforcement learning (RL)~\cite{peng2018deepmimic,ho2016generative,merel2020catch} and advanced modeling techniques (\eg, MoE~\cite{won2020scalable,hua2021learning,chen2022towards}, VAE~\cite{merel2018neural,lee2020stochastic}, and GAN~\cite{ho2016generative,song2018multi}) alongside CLIP features~\cite{rocamonde2023vision,kumar2023words} attempts to improve generalization, yet faces the grand challenge of achieving physical plausibility in open vocabulary. Our method combines a shared low-level controller with a high-level policy tailored to each instruction, ensuring actions are physically realistic and adaptable to diverse instructions.

\textbf{Open-vocabulary motion generation} creates human motions from natural language descriptions outside the training distribution. Leveraging large-scale motion-language datasets~\cite{punnakkal2021babel,guo2022generating,lin2023motion}, generative models have shown promise in motion synthesis~\cite{tevet2022human,zhang2023generating,ren2023diffusion,zhang2023motiongpt,jiang2024motiongpt}. However, these models often struggle with zero-shot generalization or adhering to the laws of physics, limited by their training data scope. Attempts to address these limitations include simplifying complex instructions with Large Language Models~\cite{kalakonda2022action,kumar2023words} and employing pretrained \acp{vlm} like CLIP for supervision~\cite{lin2023being,tevet2022motionclip,hong2022avatarclip}, yet achieving natural and physics-compliant motions remains a significant hurdle. Our method builds upon these foundations, seeking to generate interactive and physically plausible motions from open-vocabulary descriptions, distinguishing itself from approaches like \ac{vlm}-RMs~\cite{rocamonde2023vision} by modeling motion priors more effectively.

\textbf{Humanoid object interaction}, a relatively uncharted territory in physics-based motion generation, has seen simplifications such as attaching objects to characters' hands to bypass the complexity of modeling physical interactions~\cite{peng2018deepmimic,zhang2023learning,xu2023composite}. For dynamic interactions, encoding object states (positions and velocities) into the agent's observations has facilitated specific tasks like dribbling~\cite{peng2021amp,peng2019mcp} and interacting with furniture~\cite{hassan2023synthesizing}, albeit requiring precise, object-specific rewards. This state-based approach is less feasible in open environments with diverse objects. Alternatively, vision-based policies~\cite{merel2020catch} have shown potential for broader applications but are limited by their training domains. Our approach leverages a \ac{vlm} for a more generalized motion-text alignment, avoiding the intricacies of manual reward crafting for varied interactive tasks.

\section{\textbf{\model}}\label{sec:method}

\begin{figure}[t!]
    \centering	
    \includegraphics[width=\linewidth]{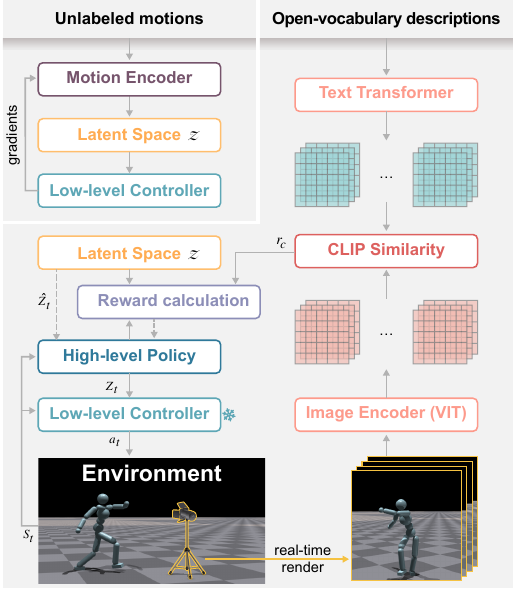}
    \caption{\textbf{The hierarchical structure of \model{}.} Initially, the low-level controller (top-left) is trained to encode unlabeled motions into a shared latent space $\mathcal{Z}$. Subsequently, for each open-vocabulary text description, a high-level policy is trained. This policy orchestrates low-level actions to optimize the CLIP similarity between rendered images and the provided text, effectively composing actions that align with the textual instructions.}
    \label{fig:model}
\end{figure}

\model{} consists of two core components: the \textbf{low-level controller} and the \textbf{high-level policy}, illustrated in \cref{fig:model}. Initially, we train a shared low-level controller, $\pi^L$, using unlabeled motion clips to distill a latent representation of atomic actions. This process utilizes \ac{gail}~\cite{ho2016generative}, guaranteeing that the atomic actions are physically plausible.

Subsequently, for each open-vocabulary textual instruction, we train a high-level policy, $\pi^H$, tasked with composing atomic actions derived from low-level controllers. This high-level policy leverages a \textbf{flexible and generalizable image-based reward} via a \ac{vlm}. This design facilitates the learning of physical interactions with dynamic objects, obviating the need for handcrafted reward engineering.

\subsection{Low-Level Controller}

The low-level controller, inspired by CALM~\cite{tessler2023calm}, enables the physically simulated humanoid agent to learn a diverse set of atomic actions. Formally, given an unlabeled motion dataset $\mathcal{M}$, we simultaneously train a motion encoder $E$, a discriminator $D$, and a controller $\pi^L(a|s, z)$. Here, $a$ denotes the action, $s$ the state, and $z \in \mathcal{Z}$ the latent motion representation. The state $s$ comprises the agent's current root position, orientation, joint positions, and velocities, while the action $a$ specifies the next target joint rotations.

Training proceeds as follows: A motion clip $M$ from $\mathcal{M}$ is encoded by $E$ to yield the latent representation $z=E(M)$. The controller $\pi^L(a|s,z)$ generates an action $a$ based on the current state $s$ and latent $z$. The agent then executes the action $a$ in the physics-based simulator with a PD controller, resulting in a new state $s'$. 

The discriminator $D$ distinguishes whether the given $(s, s')$ originates from the motion $M$ corresponding to $z$, is produced by the controller $\pi^L$ following the latent code $z$, or is produced by $\pi^L$ following another latent code $z'\sim\mathcal{Z}$. We train $D$ with a ternary adversarial loss:
\begin{align*}
    \small
    &\mathcal{L}_\mathcal{D} = - \mathbb{E}_{M \in \mathcal{M}} \Big( \mathbb{E}_{d^\pi (s, s' | z)} \left[ \log \left( 1 - \mathcal{D} (s, s' | z) \right) \right] \numberthis \\
    &+ \mathbb{E}_{d^M (s, s')} \left[ \log \mathcal{D} (s, s' | z) + \log \left( 1 - \mathcal{D} (s, s' | z' \sim \mathcal{Z}) \right) \right] \\
    &+  w_\text{gp} \mathbb{E}_{d^\mathcal{M} (s, s')} \left[ || \nabla_\theta \mathcal{D}(\theta) |_{\theta = (s, s' | \hat{z})} ||^2 \right] \Big | \hat{z}=\text{sg}(E(M)) \Big),
\end{align*}
incorporating a gradient penalty with coefficient $w_{\text{gp}}$ for stability, where $\text{sg}(\cdot)$ denotes the stop gradient operator.

The encoder $E$ is refined with both alignment and uniformity losses to ensure that embeddings of similar motions are closely aligned in the latent space, while dissimilar ones remain distinct~\cite{wang2020understanding}, thus structuring $\mathcal{Z}$ effectively.

The controller $\pi^L$ aims to maximize the GAIL reward from $D$, calculated as
\begin{equation}
    \small
    r^L(s, s', z) = - \log \left(1 - \mathcal{D}\left(s, s' | z \right) \right),
\end{equation}
encouraging the generation of motions that closely resemble the original motion $M$ associated with latent code $z$.

\subsection{High-Level Policy}

Building upon the atomic action repository created by the low-level controller, the high-level policy's objective is to compose these actions, via the control of latent representation $z$, to generate motions that align with given text descriptions. With the low-level controller $\pi^L$ fixed, we train a high-level policy $\pi^H$ for each specific textual instruction, ensuring that the combined operation of both policy levels results in motions congruent with the text. The training process for the high-level policy is outlined in \cref{alg:high}.

\begin{algorithm}[ht!]
    \small
    \SetAlgoLined 
    \caption{Training of the high-level policy}
         \label{alg:high}
    \KwIn{Reference motion dataset $\mathcal{M}$, frozen low-level controller $\pi^L$, frozen motion encoder $E$, simulation environment \textsc{env}, renderer image $\mathcal{I}$, CLIP feature of the description text $f_d$}
    $\mathcal{Z}$ = $E(\mathcal{M})$ initialize motion latent space \\
    \While{not converged}{
        $\mathcal{B}\leftarrow\emptyset$; $p \leftarrow 0$ initialize \\
        \For{horzion\_length $= 1,...,n$}{
            sample $\hat{z}$ from $\mathcal{Z}$ \\
            \eIf{horzion\_length $= 1$}{
                    $s \leftarrow \mathrm{initialize}$;
                    $z \leftarrow \hat{z}$ \\
                }{
                    $s \leftarrow \textsc{env}(s,a)$; 
                    $z \leftarrow \pi^H(s)$\\
                }
            \For{llc\_steps $= 1,...,t$}{
                $s\leftarrow\textsc{env}(s,\pi^L(s,z))$ step simulation\\
                $r^H$ $\leftarrow$ calculate reward with \cref{eq:clip}\\
                \If{$\textsc{head\_height} < 0.15$}{
                    $s,p\leftarrow0$ reset agent and counter\\
                }
                \If{$\mathrm{similarity\ is\ less\ than\ last\ step}$} {
                    $p\leftarrow p+1$ increment counter \\
                    \If{$p \geq 8$} {
                        $p\leftarrow0$ reset counter \\
                        reset $s$ with 80\% probability
                    }
                }
                
            }
            update $\mathcal{B}$ and $\pi^H$ according to PPO\\
        }
    }
\end{algorithm}

The high-level policy $\pi^H$ is implemented as an MLP, taking the agent's state $s$ as input and outputting a latent representation $z$ close to the low-level controller's latent space $\mathcal{Z}$. It is trained using a composite reward of image-based similarity and latent-representation alignment. Given state $s$ and text description $d$, we render the agent's image $\mathcal{I}(s)$ and encode it along with the text using a pretrained, frozen CLIP model to obtain features $f_\mathcal{I}$ and $f_d$. The similarity reward is computed as the cosine similarity between $f_\mathcal{I}$ and $f_d$, with an additional latent-representation alignment reward to draw $z$ nearer to the latent distribution of $\mathcal{M}$. The combined reward is given by:
\begin{equation}
    \small
    r^H = \omega_c\cdot\frac{f_\mathcal{I}\cdot f_d}{\vert f_\mathcal{I}\vert \vert f_d \vert} + \omega_s\cdot\text{exp}(-4\Vert z - \hat{z} \Vert_2),
    \label{eq:clip}
\end{equation}
where $\omega_c, \omega_s$ are weighting factors, and $\hat{z}$ is a sample from $\mathcal{Z}$. This image-based reward mechanism enables \model to achieve text-to-motion alignment for open-vocabulary instructions. In addition, the image-based representation naturally encodes the entire environment around the agent, thus facilitating object interactions without modifying the encoding or architecture.

\subsection{Implementation Details}

\paragraph{Low-level controller}

The architecture of the encoder, low-level control policy, and discriminator comprises MLPs with hidden layers sized [1024, 1024, 512]. The latent space $\mathcal{Z}$ is 64-dimensional. The alignment loss is set to 0.1, uniformity loss to 0.05, and gradient penalty to 5. The low-level controller is optimized using PPO~\cite{schulman2017proximal} in IsaacGym. The training process is conducted on a single A100 GPU, operating at a 120Hz simulation frequency, and spans four days to cover a dataset comprising 93 unique motion patterns. Detailed hyperparameter settings of the low-level controller can be found in \cref{supp:tab:low}.

\paragraph{High-level policy}

The high-level policy, implemented as a two-layer MLP with hidden units of [1024, 512], outputs a 64-dimensional vector and is optimized using PPO. Training is conducted on an NVIDIA RTX3090 GPU, taking approximately 2.2 hours. Operationally, the high-level policy executes at a frequency of 6Hz, in contrast to the low-level policy, which operates at a more rapid 30Hz. This discrepancy in execution rates is strategic; the high-level policy is invoked every five timesteps, granting the low-level controller sufficient time to act on a given stable latent representation $z$ and execute a complete atomic action. Such a setup is crucial for preventing the emergence of unnatural motion sequences by ensuring that each selected atomic action is fully realized before transitioning. Detailed hyperparameters of the high-level policy can be found in \cref{supp:tab:high}.

To further refine the training process and motion quality, an early termination strategy is employed to circumvent potential pitfalls of the high-level policy becoming trapped in suboptimal local minima. Specifically, the environment is reset with an 80\% probability following eight successive reductions in CLIP similarity, or deterministically if the agent's head height falls below 15cm. This approach significantly enhances training efficiency and the fidelity of the generated motions, ensuring a balance between exploration and the avoidance of poor performance traps. 

\paragraph{Rendering}

We use IsaacGym's default renderer, positioning the camera at (3m, 0m, 1m) while the agent is initialized at the origin. To maintain the agent at the focus of our visual feedback, we dynamically adjust the camera's orientation each timestep to align with the agent's pelvis joint. To encode the rendered images into a feature space compatible with our learning objectives, we employ the \texttt{CLIP-ViT-B/32} model checkpoint from OpenCLIP~\cite{cherti2023reproducible}, leveraging its robust representational capabilities.

\paragraph{State projection}

Given the computational demands of rendering images and extracting their CLIP features, we streamline the training process by introducing an MLP that projects the agent's state vectors $s$ directly to CLIP image features. This projection MLP is fine-tuned with an MSE loss against 104 million agent states accumulated during the high-level policy training. By substituting the render-and-encode steps with this MLP, we achieve a significant speedup, enhancing training efficiency by approximately 10.4 times, thereby mitigating the bottleneck associated with real-time image rendering and feature extraction.

\section{Experiments}\label{sec:exp}

In this section, we detail the motion dataset curation for \model's low-level controller training (\cref{sec:exp:low}), evaluate \model's open-vocabulary motion generation against others (\cref{sec:exp:high}), analyze the text enhancement impact on effectiveness (\cref{sec:exp:text}), showcase physical interaction examples (\cref{sec:exp:interact}), and compare our reward design with existing formulations (\cref{sec:exp:reward}).

\subsection{Training of Low-Level Controller}\label{sec:exp:low}

\paragraph{Dataset}

To enrich the low-level controller with diverse atomic actions, we assembled a dataset of 93 distinct motion records, primarily sourced from the CMU Graphics Lab Motion Capture Database~\cite{cmu_mocap} and SFU Motion Capture Database~\cite{sfu_mocap}. This collection spans various action categories, including locomotion (\eg, walking, running, jumping), dance (\eg, jazz, ballet), acrobatics (\eg, roundhouse kicks), and interactive gestures (\eg, pushing, greeting), all retargeted to a humanoid skeleton with 15 bones. We also adjusted any motions that lacked physical plausibility, ensuring the dataset's fidelity for effective imitation learning.

\begin{figure}[t!]
    \centering	
    \includegraphics[width=\linewidth]{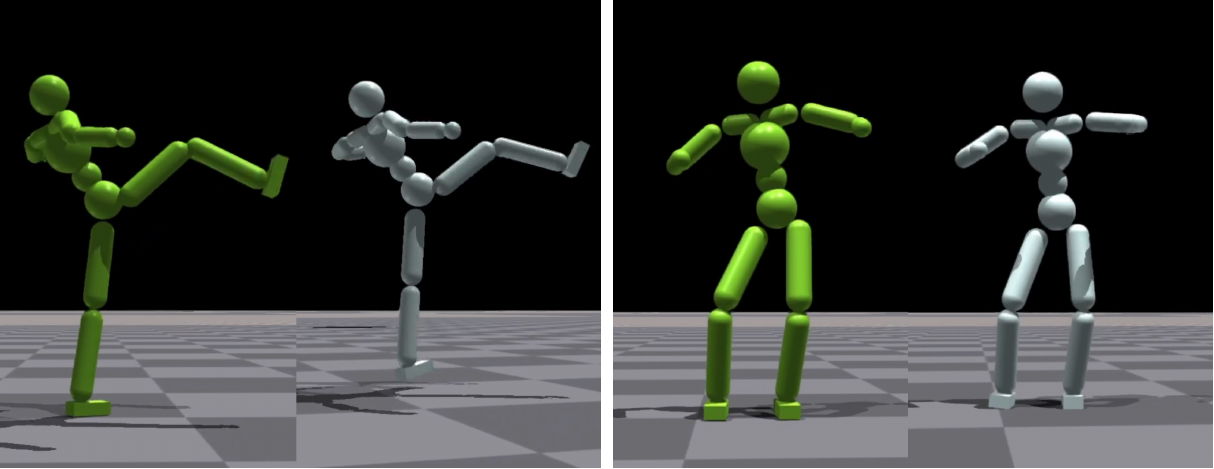}
    \caption{\textbf{Atomic actions from the trained low-level controller.} Each subfigure depicts the green agent demonstrating the reference motion from the dataset, while the white agent illustrates the corresponding learned atomic action.}
    \label{fig:low}
\end{figure}

\paragraph{Training stabilization}

Adversarial imitation learning's instability, influenced by the volume and distribution of training data, can skew the density distribution in latent space, limiting the diversity of atomic actions for high-level policy selection. To mitigate this, we categorized motion records into 3 primary and 4 secondary groups by action scale and involved limbs. Details of the category division are described in \cref{supp:motiondata}. By adjusting training data weights, we increased the likelihood of less frequent action groups, ensuring the variety of learned atomic actions; see also \cref{fig:low} and \cref{supp:fig:low}.

\begin{figure}[b!]
    \centering
    \includegraphics[width=\linewidth]{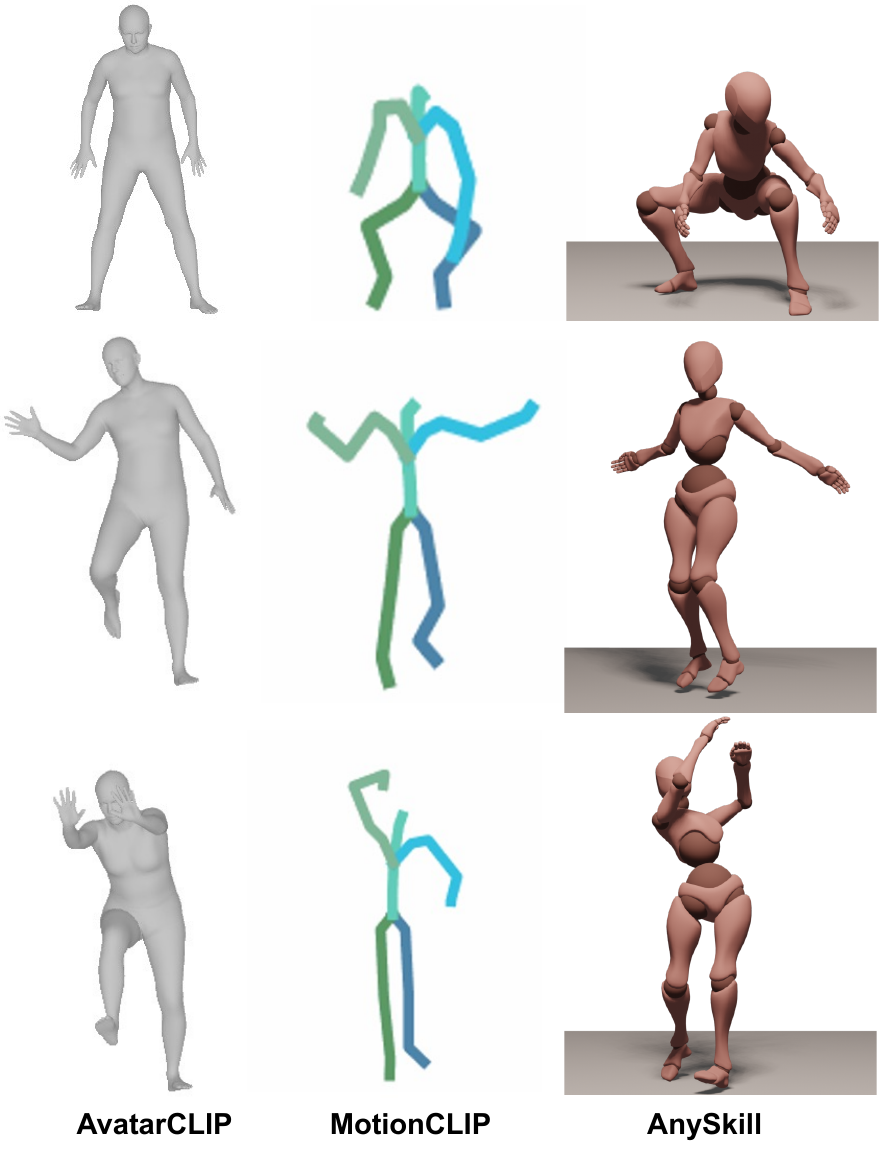}
    \caption{\textbf{Qualitative comparisons on open-vocabulary motion generation.} From top to bottom, the descriptions are \emph{``sit down, bent torso, legs folded at knees''}, \emph{``legs off the ground, wave hands''}, and \emph{``coiling the arm, throw a ball''}. We showcase the most representative frames that best align with the descriptions.}
    \label{fig:motion_data}
\end{figure}

\begin{figure*}[t!]
    \centering
    \begin{subfigure}[b]{\linewidth}
        \includegraphics[width=\linewidth]{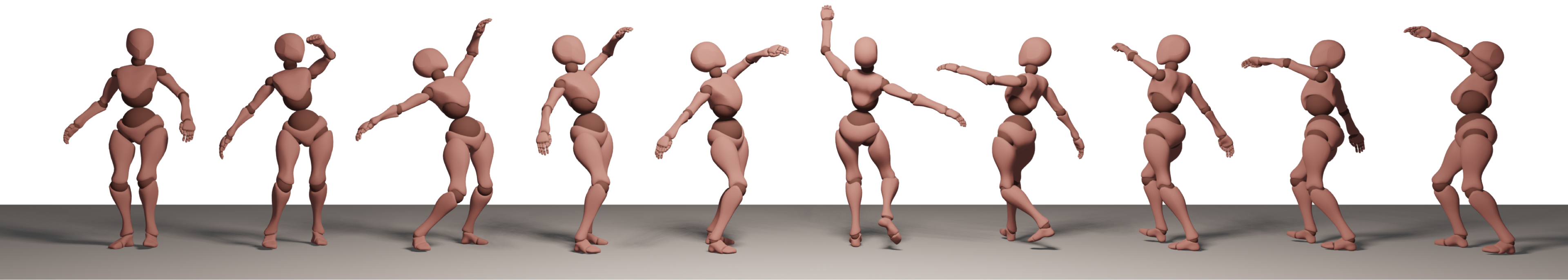}
        \caption{dance and turn around}
        \label{fig:sub1}
    \end{subfigure}%
    \\%
    \begin{subfigure}[b]{\linewidth}
        \includegraphics[width=\linewidth]{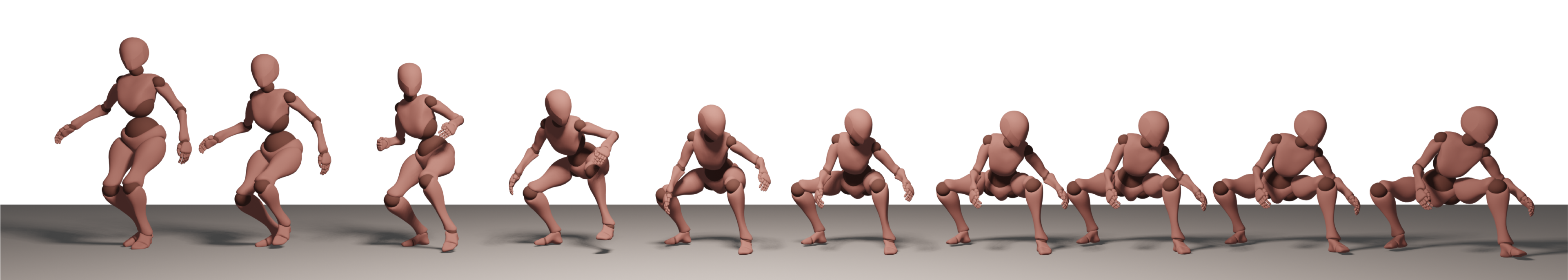}
        \caption{sit down, bent torso, legs folded at knees}
        \label{fig:sub2}
    \end{subfigure}%
    \\%
    \begin{subfigure}[b]{\linewidth}
        \includegraphics[width=\linewidth]{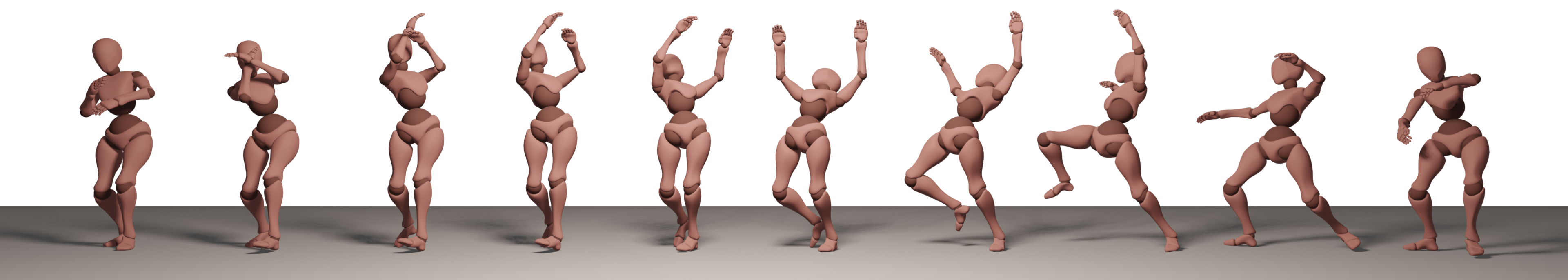}
        \caption{coiling the arm, throw a ball}
        \label{fig:sub3}
    \end{subfigure}%
    \\%
    \begin{subfigure}[b]{\linewidth}
        \includegraphics[width=\linewidth]{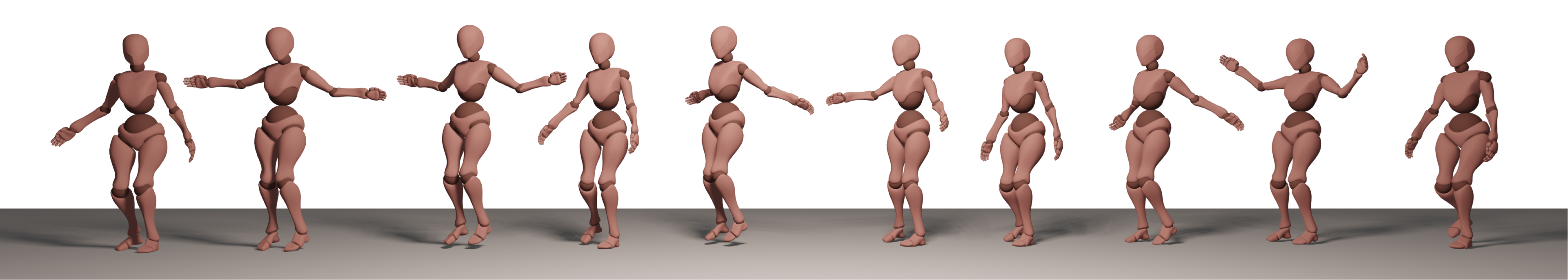}
        \caption{legs off the ground, wave hands}
        \label{fig:sub4}
    \end{subfigure}%
    \\%
    \begin{subfigure}[b]{\linewidth}
        \includegraphics[width=\linewidth]{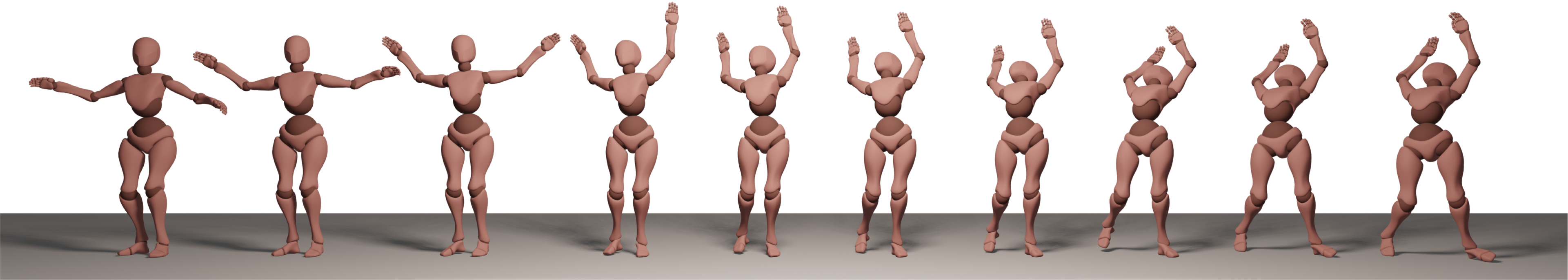}
        \caption{raise two arms}
        \label{fig:sub5}
    \end{subfigure}%
    \caption{\textbf{Qualitative results of generated motion by \model.} Displayed are specific text descriptions and the corresponding motions generated by \model, as evaluated in the user study. Motion sequences progress from left to right.}
    \label{fig:high}
\end{figure*}

\subsection{\textbf{\model{}} Evaluation}\label{sec:exp:high}

Given the nascent field of open-vocabulary physical skill learning, we benchmark \model against the two foremost similar methods in open-vocabulary motion generation: MotionCLIP~\cite{tevet2022motionclip} and AvatarCLIP~\cite{hong2022avatarclip}, which also utilize CLIP similarity for generating human motions. To further understand the efficacy of our approach, we introduce a variant of our method, ``Ours (no ET),'' which operates without the early termination strategy.

For this evaluation, we selected 5 open-vocabulary text descriptions requiring comprehensive body movement and not covered in \model's training data. To assess the generated motions, we engaged 24 MTurk workers to rate them on task completion, smoothness, naturalness, and physical plausibility, using a scale from 0 to 10. Moreover, we computed the CLIP similarity score between the rendered images and the text descriptions for each method as an objective measure. The motions generated by each method, including qualitative comparisons, are showcased in \cref{fig:motion_data}, with an in-depth look at \model's outputs presented in \cref{fig:high}. Beyond the five actions presented, additional actions are shown in \cref{supp:fig:skill}

\begin{table}[ht!]
    \centering
    \small
    \caption{\textbf{Quantitative evaluation of high-level policy.}}
    \label{tab:com}
    \setlength{\tabcolsep}{3pt}
    \resizebox{\columnwidth}{!}{%
        \begin{tabular}{llllll}
            \toprule
            \textbf{}     & \textbf{Success} $\uparrow$      & \textbf{Natural} $\uparrow$ & \textbf{Smooth}$\uparrow$ & \textbf{Physics}$\uparrow$  & \textbf{CLIP\_S}$\uparrow$ \\
            \midrule
            AvatarCLIP~\cite{hong2022avatarclip}      &  4.29  &   4.74    &   5.79    &  5.74    &    21.11    \\
            MotionCLIP~\cite{tevet2022motionclip} &   3.16   &   4.93    &    5.72    &    5.83    &     21.16     \\ \midrule
            Ours (w/o ET)   &   5.05   &   4.88    &   5.68   &  5.31     &    21.89    \\
            Ours (w/o text-enhance)   &   3.06   &   4.48    &   5.19   &  5.96     &    20.76    \\
            Ours (w/ VideoCLIP~\cite{xu2021videoclip})   &   2.37   &   4.90    &   5.65   &  6.41     &    21.35    \\
            \textbf{Ours (full)}      &   \textbf{6.16}    &   \textbf{6.23}  & \textbf{6.51} & \textbf{6.93}  &   \textbf{24.18}            \\
            \bottomrule
        \end{tabular}%
    }%
\end{table}

We present the results of the human study and quantitative metrics in \cref{tab:com}, demonstrating that \model significantly surpasses current methods across all evaluated metrics. The ablation study underscores the importance of incorporating early termination into the training process. For additional comparative and qualitative results, see \cref{supp:fig:reward5}.

\subsection{Text Enhancement}\label{sec:exp:text}

\model excels at open-vocabulary skill acquisition, outperforming existing models. Its performance, however, is contingent on the specificity and scope of text descriptions. Performance drops with vague descriptions or for tasks requiring prolonged execution due to reliance on image-based similarity for rewards. For example, ``do yoga'' encompasses a broad range of poses, complicating convergence on a specific action. Similarly, for extended actions like ``walk in a circle,'' the model may not fully complete the task, as image-based rewards provide insufficient directional guidance. 

\begin{figure}[b!]
    \centering
    \begin{subfigure}[b]{0.333\linewidth}
        \includegraphics[width=\linewidth]{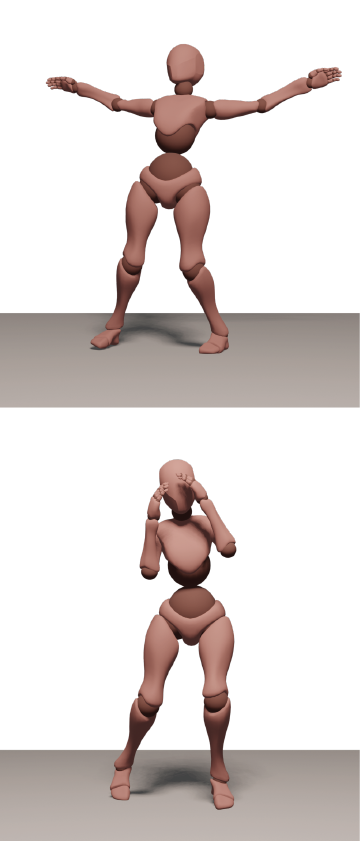}
        \caption{}
        \label{fig:wave}
    \end{subfigure}%
    \hfill%
    \begin{subfigure}[b]{0.333\linewidth}
        \includegraphics[width=\linewidth]{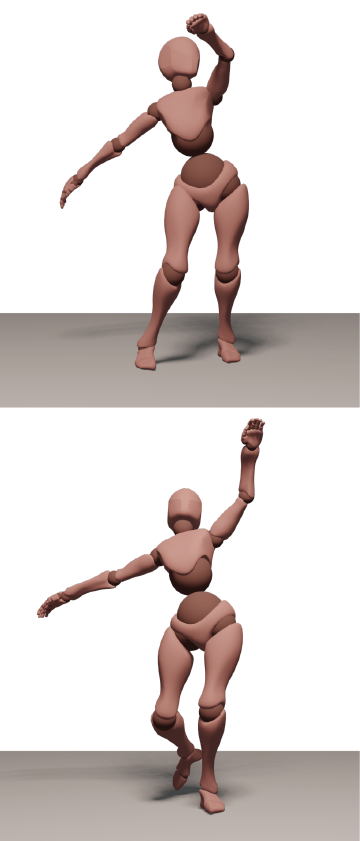}
        \caption{}
        \label{fig:dance}
    \end{subfigure}%
    \hfill%
    \begin{subfigure}[b]{0.333\linewidth}
        \includegraphics[width=\linewidth]{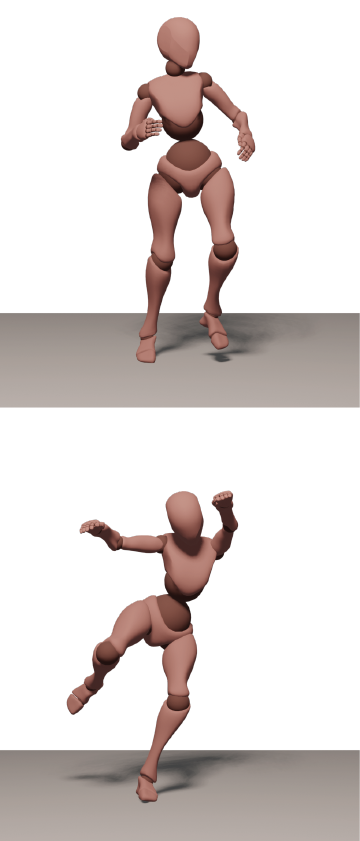}
        \caption{}
        \label{fig:kick}
    \end{subfigure}%
    \caption{\textbf{Qualitative evaluation of text description enhancement.} We compare motions generated with original HumanML3D~\cite{guo2022generating} descriptions (top row) against those from our enhanced descriptions (bottom row). Text descriptions are (a) \emph{``wave hi'' and ``raised arm bent at the elbow''}; (b) \emph{``Waltz dance'' and ``left foot step backward, right hand extends''}; (c) \emph{``kick'' and ``left leg forward, right leg retreats''}.}
    \label{fig:motion_txt}
\end{figure}

To counteract these limitations, we introduced an automated script utilizing GPT-4~\cite{openai2023gpt4} to refine and clarify textual instructions, enhancing specificity and reducing potential motion interpretation ambiguity. This refinement process significantly improves \model's execution accuracy. \cref{fig:motion_txt} compares the original and refined texts alongside their generated motions; see \cref{supp:exp:text} for more qualitative results.

Moreover, we refined text descriptions from the HumanML3D~\cite{guo2022generating} and BABEL~\cite{punnakkal2021babel} databases, amassing 1,896 unique, enhanced text instructions. For comprehensive details on the refined texts and their impact on motion generation, refer to \cref{supp:textdata}.

\subsection{Interaction Motions}\label{sec:exp:interact}

\begin{figure}[t!]
    \centering
    \includegraphics[width=\linewidth]{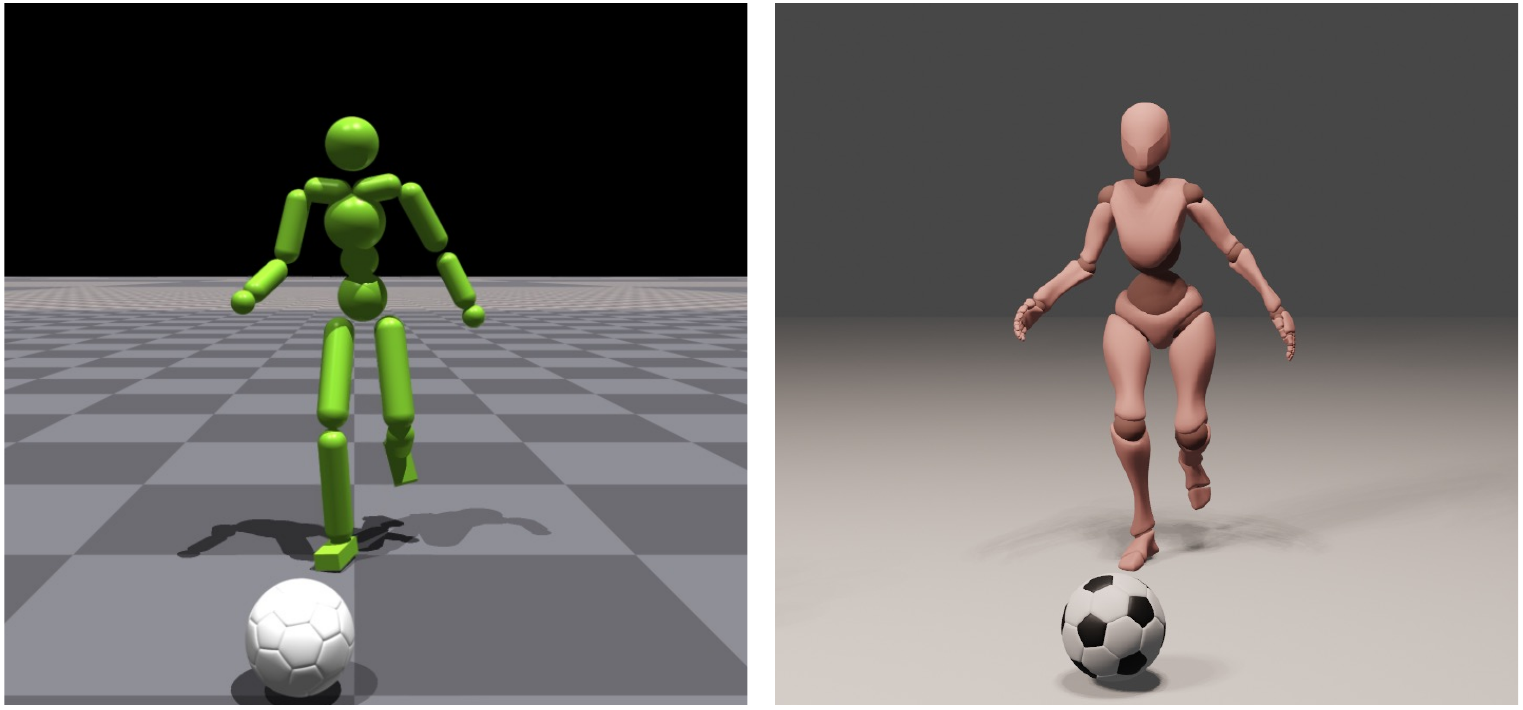}
    \caption{\textbf{Agent and rendered mesh.} The simulation of our agent and the interacting object (left) alongside their visualization (right).}
    \label{fig:sim}
\end{figure}

\begin{figure*}[t!]
    \centering
    \begin{subfigure}[b]{\linewidth}
        \includegraphics[width=\linewidth]{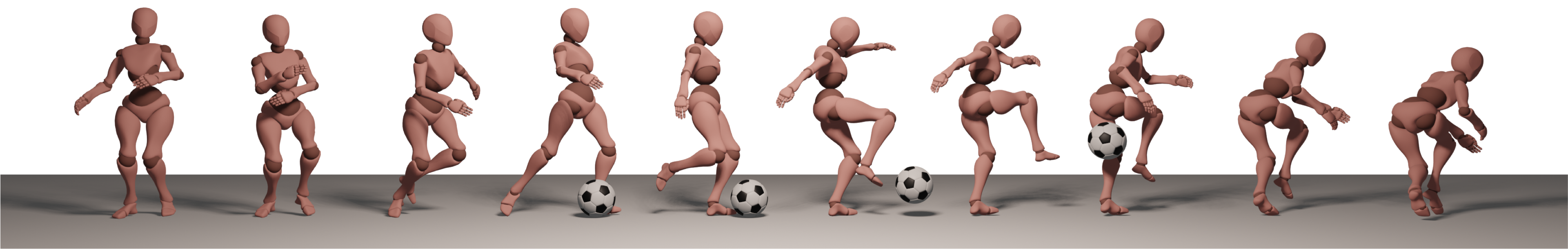}
        \caption{kick the white ball}
        \label{fig:soccer1}
    \end{subfigure}%
    \\%
    \begin{subfigure}[b]{\linewidth}
        \includegraphics[width=\linewidth]{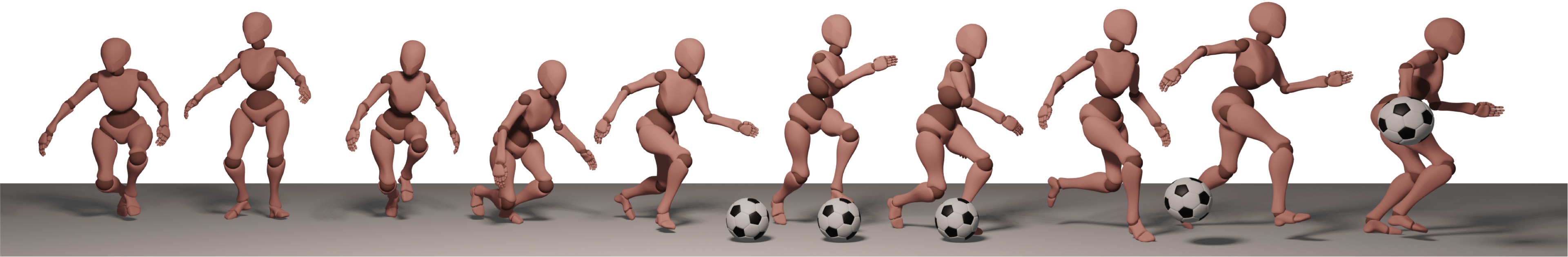}
        \caption{move the white ball}
        \label{fig:soccer2}
    \end{subfigure}%
    \\%
    \begin{subfigure}[b]{\linewidth}
        \includegraphics[width=\linewidth]{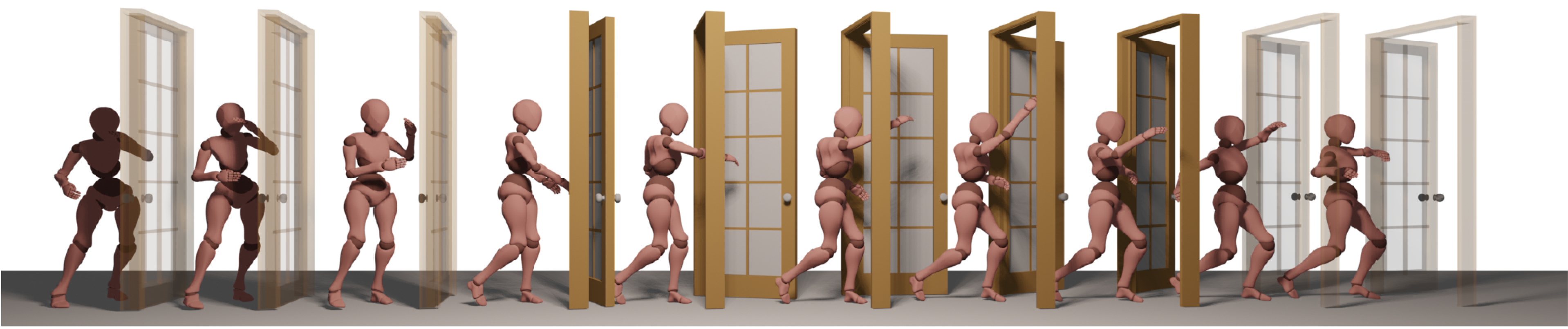}
        \caption{raise arm, open the door}
        \label{fig:door}
    \end{subfigure}%
    \caption{\textbf{Interaction motions generated by \model.} Displayed are interaction sequences by \model: two with a soccer ball (a-b) and one with a door (c), progressing from left to right.}
    \label{fig:obj}
\end{figure*}

\model demonstrates the superb capability to interact with dynamic objects, for instance, a \emph{soccer ball} and a \emph{door}. To capture these interactions accurately during training, we manually adjust the camera positions, focusing on the door and soccer ball. The alignment between the simulation environment and the rendered visualizations is showcased in \cref{fig:sim}. The qualitative assessments, as seen in \cref{fig:obj}, along with the quantitative evaluations in \cref{tab:inter}, confirm that \model efficiently learns to interact with a variety of objects without necessitating any modifications to its learning algorithm or reward design. Our tests primarily involve interactions with a single object, yet extending \model to engage with multiple objects concurrently is anticipated to be straightforward. Further interactive motions with various objects are available in \cref{supp:interact} and \cref{supp:fig:scene}.

\begin{table}[ht!]
    \centering
    \small
    \caption{\textbf{Quantitative evaluation of interaction motions.}}
    \label{tab:inter}
    \setlength{\tabcolsep}{3pt}
    \resizebox{\columnwidth}{!}{%
        \begin{tabular}{llllll}
        \toprule
        \textbf{}     & \textbf{Success} $\uparrow$      & \textbf{Natural} $\uparrow$ & \textbf{Smooth}$\uparrow$ & \textbf{Physics}$\uparrow$  & \textbf{CLIP\_S}$\uparrow$ \\
        \midrule
        Interaction w. object &  5.42 \footnotesize{\gr{-0.74}}  &   5.62 \footnotesize{\gr{-0.61}}   &   5.34 \footnotesize{\gr{-1.17}}    &  5.45 \footnotesize{\gr{-1.48}}   &  24.49 \footnotesize{\rr{+0.35}}\\
        Interaction w. scene &   4.53 \footnotesize{\gr{-1.63}}  &   4.47 \footnotesize{\gr{-1.76}}   &    5.01 \footnotesize{\gr{-1.50}}  &  5.41 \footnotesize{\gr{-1.52}}  &   22.41 \footnotesize{\gr{-1.73}}\\
        \bottomrule
        \end{tabular}%
    }%
\end{table}

\subsection{Reward Function Analysis}\label{sec:exp:reward}

We evaluate 4 recent reward functions image- and physics-based RL and compare them with ours using cosine similarity. These include \ac{vlm}-RMs~\cite{rocamonde2023vision}, which adjusts the CLIP feature of text to exclude agent-specific details; CLIP-S~\cite{zhao2024testtime}, applying a modified CLIP similarity as the reward; VideoCLIP~\cite{xu2021videoclip}, calculating mean-pooled CLIP features across frames for temporal coherence; and ASE~\cite{peng2022ase}, adding a velocity reward for desired agent movement.

Using these rewards, we train \model on identical descriptions and assess motion quality via a user study similar to the one described in \cref{sec:exp:high}, with results presented in \cref{tab:abla,supp:exp:reward}. Our approach surpasses the baseline methods in most metrics, demonstrating the effectiveness of our reward function. Notably, AvgPool scores highly in smoothness, benefiting from averaging alignment scores over time.

\begin{table}[ht!]
    \centering
    \small
    \caption{\textbf{Comparisons of the reward design.}}
    \label{tab:abla}
    \setlength{\tabcolsep}{3pt}
    \resizebox{\columnwidth}{!}{%
        \begin{tabular}{llllll}
            \toprule
            \textbf{}        & \textbf{Success}$\uparrow$    & \textbf{Natural}$\uparrow$ & \textbf{Smooth}$\uparrow$ & \textbf{Physics}$\uparrow$ & \textbf{CLIP\_S}$\uparrow$ \\
            \midrule
            \ac{vlm}-RMs~\cite{rocamonde2023vision}     &    3.15    &    4.36     &   5.35   &   5.17  &      19.46     \\
            CLIP-S~\cite{zhao2024testtime}      &   3.80  &   5.41  &    5.98 &   6.21  &     19.78    \\
            AvgPool~\cite{xu2021videoclip}     &   5.09   &  5.96   &  \textbf{6.55}   &  6.70    &        20.25          \\
            + vel. rew.~\cite{peng2022ase} &    2.73   &     4.42     &   5.35    &   5.22   &   18.39   \\
            \midrule
            \textbf{Ours}      &   \textbf{6.16}    &   \textbf{6.23}  &  6.51   & \textbf{6.93}  &  \textbf{24.18}  \\
        \bottomrule
    \end{tabular}%
}%
\end{table}

\section{Conclusion}

We introduced \model, a novel hierarchical framework for acquiring open-vocabulary physical interaction skills, combining an imitation-based low-level controller for motion generation with a robust, flexible image-based reward mechanism for adaptable skill learning. Through qualitative and quantitative assessments, \model is the first method capable of extending learning to encompass unseen tasks and interactions with novel objects, opening new venues in motion generation for interactive virtual agents.

\paragraph{Future directions}

\model's potential and limitations are closely linked to the CLIP model's capabilities, guiding its current success and defining its challenges. As noted in \cref{sec:exp:text}, reliance on image-based rewards restricts \model's effectiveness in scenarios with prolonged durations or visual ambiguity. Future work aims to address these issues by enhancing the model's understanding of temporal dynamics, integrating sophisticated multimodal alignment strategies, and incorporating interactive feedback loops.

The current need to develop a specialized policy for each new task---requiring substantial training time and resources---highlights a direction for future work: transforming \model into a more universally applicable framework. This evolution will streamline the process of skill acquisition, dramatically reducing the time and resources required to master new interactive abilities. By achieving this, we anticipate enabling \model to learn an array of skills in a unified, efficient manner, significantly broadening the scope of applications for interactive virtual agents and making sophisticated motion generation more accessible.

\paragraph{Acknowledgement}

The authors would like to thank Ms. Zhen Chen (BIGAI) for her exceptional contribution to the figure designs, Yanran Zhang and Jiale Yu (Tsinghua University) for their invaluable assistance in the experiments and prompt design, and Huiying Li (BIGAI) for crafting the agent's appearance. We also thank NVIDIA for generously providing the necessary GPUs and hardware support. This work is supported in part by the National Science and Technology Major Project (2022ZD0114900), an NSFC fund (62376009), and the Beijing Nova Program.

{
    \small
    \bibliographystyle{ieeenat_fullname}
    \bibliography{reference_header,references}
}

\clearpage
\appendix
\renewcommand\thefigure{A\arabic{figure}}
\setcounter{figure}{0}
\renewcommand\thetable{A\arabic{table}}
\setcounter{table}{0}
\renewcommand\theequation{A\arabic{equation}}
\setcounter{equation}{0}
\pagenumbering{arabic}
\renewcommand*{\thepage}{A\arabic{page}}
\setcounter{footnote}{0}

\section{Data}

This section offers a detailed account of the data's origins and the methodologies employed for its processing.

\subsection{Text Data}\label{supp:textdata}

Text descriptions sourced from publicly available online datasets are often marked by redundancy, ambiguity, and insufficient detail. To address these issues, it is necessary to preprocess the descriptions to render them more practical and usable. For generating practical text descriptions, we implemented a three-tiered process leveraging GPT-4~\cite{openai2023gpt4}. This encompasses \textbf{filtering text} to discard non-essential details, \textbf{scoring text} for assessing utility, and \textbf{rewriting text} to improve clarity and applicability. Our goal is to identify text descriptions that significantly contribute to mastering open-vocabulary physical skills from a robust pre-existing dataset, and to standardize the collection of text instructions.

\paragraph{Filter text}

Initially, we compiled 89,910 text entries from HumanML3D~\cite{guo2022generating} and Babel~\cite{punnakkal2021babel}, discovering substantial repetition, including exact duplicates, descriptions of akin actions (\eg, \emph{``A person walks down a set of stairs''} vs. \emph{``A person walks down stairs''}), frequency-related repetitions (\eg, \emph{``A person sways side to side multiple times''} vs. \emph{``A person sways from side to side''}), and semantic duplicates (\eg, \emph{``The person is doing a waltz dance''} vs. \emph{``A man waltzes backward in a circle''}).

To address this issue, we initiated a deduplication process, first eliminating descriptions that were overly brief (under three tokens) or excessively lengthy (over 77 tokens). We then utilized the \textsc{llama-2-7b model} with its 4096-dimensional embedding vector for further deduplication. By computing cosine similarities between each description pair and applying a 0.92 similarity threshold, descriptions exceeding this threshold were considered repetition. This procedure refined our dataset to 4,910 unique descriptions.

\begin{figure}[ht!]
    \colorbox{myblue}{
        \begin{tabular}{p{0.9\linewidth}}
            \rowcolor{myblue!200}\color{white}
            \large\textbf{Score Prompt I}\\[+3ex]
            \midrule 
            \rowcolor{gray!10}
            You are a language expert. Please rate the following actions on a scale of 0 to 10 based on their use of language.
            \vspace{3mm}
            The requirements are:
            \begin{enumerate}[before=\itshape,font=\normalfont]
                \item The description should be fluent and concise.
                \item The description should correspond to a single human pose, instead of a range of possible poses.
                \item The description should describe a human pose at a short sequence of frames instead of a long sequence of frames (this requirement is not mandatory).
                \item If the description contains sequential logic, rate it lower. ''Walk in a circle'' is a kind of sequential logic.
                \item Except for the subject, the description should have only one verb and one noun.
                \item If the description is vivid(like ''dances like Michael Jackson''), rate it higher.
            \end{enumerate}
            \vspace{3mm}
            Here are some examples you graded in the last round:
            \begin{itemize}[before=\itshape,font=\normalfont]
                \item 6 - A person is swimming with his arms.
                \item 3 - Sway your hips from side to side.
                \item 7 - A person smashed a tennis ball.
                \item 4 - A person is in the process of sitting down.
                \item 5 - A person brings up both hands to eye level.
                \item 9 - A person dances like Michael Jackson.
                \item 2 - A person packs food in the fridge.
                \item 5 - A person flips both arms up and down.
                \item 8 - Looks like disco dancing.
                \item 3 - Kneeling person stands up.
                \item 1 - A person does a gesture while doing kudo.
                \item 6 - A person unzipping pants flyer.
                \item 0 - then kneels on both knees on the floor.
                \item 2 - A person is playing pitch and catch.
                \item 1 - A person gesturing them walking backward.
                \item 4 - A person seems confident and aggressive.
                \item 1 - A person circles around with both arms out.
                \item 5 - A person prepares to take a long jump.
                \item 6 - A person jumps twice into the air.
                \item 0 - Turning around and walking back.
            \end{itemize}\vspace{3mm}
            
            Now, please provide your actions in the format 'x - yyyy,' where 'x' is the score, and 'yyyy' is the original sentence. Please note that Do not change the original sentence.
        \end{tabular}%
    }%
    \caption{\textbf{Score Prompt I}. This prompt focuses on filtering text descriptions for fluency, conciseness, and specificity, particularly targeting individual human poses within a short sequence of frames.}
    \label{fig:prompt:score1}
\end{figure}

\paragraph{Scoring text}

After filtering out duplicates and semantically similar actions, we encountered issues like typographical errors, overly complex descriptions, and significant ambiguities in the remaining texts. These problems rendered the descriptions unsuitable for generating actionable human motion skills despite their uniqueness.

To further refine our text instructions, we evaluated the remaining descriptions for their suitability in model processing and practical motion generation. Our evaluation, detailed in \cref{fig:prompt:score1}, focused on fluency, conciseness, and the specificity of individual human poses within a brief sequence of frames. Descriptions that were direct and descriptive, containing clear verbs and nouns, were preferred over those with a sequential or ambiguous nature. Using a standardized scoring process, we ranked the action descriptions by their scores. After addressing issues in an initial round of scoring, a second evaluation was conducted to fine-tune our selection, as mentioned in \cref{fig:prompt:score2}. This led to the exclusion of descriptions within certain score ranges (0-0.92, 0.98-0.99), resulting in a curated dataset of 1,896 unique action descriptions optimized for model training.

\begin{figure}[ht!]
    \colorbox{myblue}{
        \begin{tabular}{p{0.9\linewidth}}
            \rowcolor{myblue!200}\color{white}
            \large\textbf{Score Prompt II}\\[+3ex]
            \midrule
            \rowcolor{gray!10}
            You are a language expert. Please rate the following actions on a scale of 0 to 10 based on the ambiguity of the description. Examine whether this action description corresponds to a unique action. If the description corresponds to fewer actions, like ''wave with both arms'', rate it higher. If the description corresponds to abundant actions, like ''do yoga'', rate it lower.
            \vspace{3mm}
            \begin{itemize}[before=\itshape,font=\normalfont]
                \item 7 - grab items with their left hand.
                \item 8 - hold onto a handrail.
                \item 9 - do star jumps.
                \item 5 - arms slightly curled go from right to left.
                \item 3 - sit down on something.
                \item 9 - kick with the right foot.
                \item 7 - stand and put arms up.
                \item 9 - cover the mouth with the hand.
                \item 8 - stand and salute someone.
                \item 2 - break dance.
                \item 6 - spin body very fast.
                \item 7 - open bottle and drink it.
                \item 2 - do the cha-cha.
                \item 5 - do sit-ups.
                \item 4 - slowly stretch.
                \item 6 - cross a high obstacle.
                \item 7 - grab something and shake it.
                \item 4 - lift weights to get buff.
                \item 8 - move left hand upward.
                \item 7 - walk forward swiftly.
            \end{itemize}
            \vspace{3mm}
            Now, please provide your actions in the format 'x - yyyy,' where 'x' is the score, and 'yyyy' is the original sentence. Please note that Do not change the original sentence.
        \end{tabular}
    }
    \caption{\textbf{Score Prompt II}. This prompt selects for direct and richly detailed action descriptions, prioritizing clarity with a distinct verb and noun over descriptions based on sequential or complex logic.}
    \label{fig:prompt:score2}
\end{figure}

\paragraph{Rewrite text}

In the final refinement phase, we address the specificity of action descriptions, crucial for accurately generating motions. Vague descriptions, such as \emph{'jump rope'}, can lead to ambiguous interpretations and various motion realizations, challenging the model's training due to the similarity of rewards for different motions. This observation is consistent with other motion generation studies utilizing CLIP~\cite{tevet2022motionclip,hong2022avatarclip}.

To enhance the clarity and effectiveness of the reward calculation, we rephrase and detail the descriptions. For instance, \emph{'jump rope'} is clarified to \emph{'swinging a rope around your body'}, with further details like \emph{'Raise both hands and shake them continuously while simultaneously jumping up with both feet, repeating this cycle'}. Additionally, we break down actions into more discrete moments, such as \emph{'legs off the ground, wave hand'}, to improve the reward function's precision. Our methodology for this textual refinement is detailed in \cref{prompt:rewrite}.

\begin{figure}[ht!]
    \colorbox{myblue}{
        \begin{tabular}{p{0.9\linewidth}}
            \rowcolor{myblue!200}\color{white}
            \large\textbf{Rewrite Prompt}\\[+3ex]
            \midrule
            \rowcolor{gray!10}
            Describe an action of {instruction} for a humanoid agent. The description must satisfy the following conditions:
            \vspace{3mm}
            \begin{enumerate}[before=\itshape,font=\normalfont]
                \item The description should be concise.
                \item The description should describe a human pose in a single frame instead of a sequence of frames.
                \item The description should correspond to only one human pose, instead of a range of possible poses, minimize ambiguity.
                \item The description should be less than 8 words.
                \item The description should not contain a subject like ''An agent'', ''A human''.
                \item The description should have less than two verbs and two nouns.
                \item The description should not have any adjectives, adverbs, or any similar words like ''with respect''.
                \item The description should not include details describing expressions or fingers and toes.
            \end{enumerate}
            \vspace{3mm}
            For example, it's better to describe ``take a bow'' as ``bow at a right angle.''
        \end{tabular}
    }
    \caption{\textbf{Rewrite Prompt}. This prompt is designed for rephrasing action descriptions to enhance clarity and incorporate additional details, aiming to improve the specificity and effectiveness of the generated motions.}
    \label{prompt:rewrite}
\end{figure}

\subsection{Motion Data}\label{supp:motiondata}

\begin{figure*}[t!]
    \centering
    \includegraphics[width=.65\linewidth]{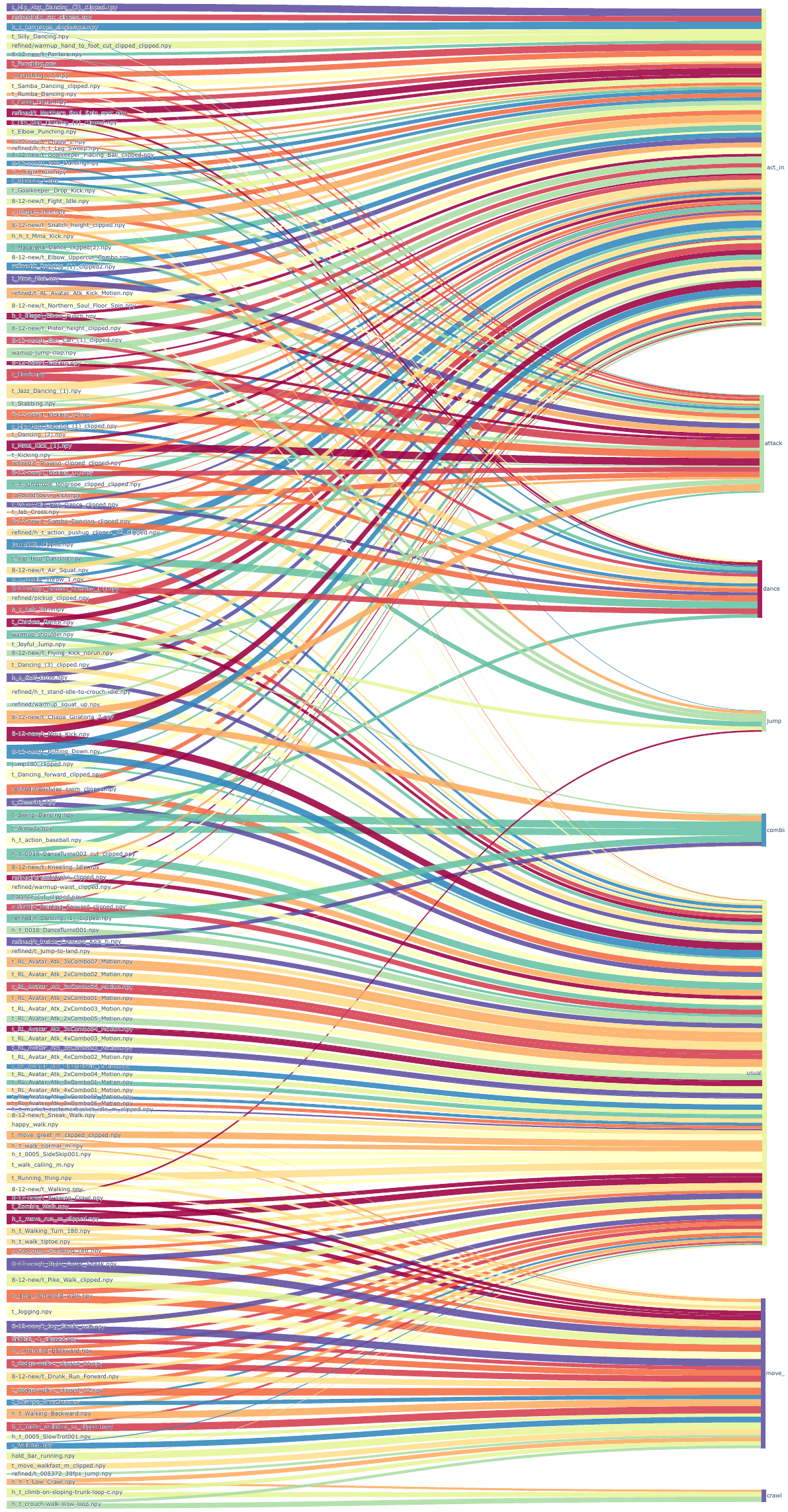}
    \caption{\textbf{The distribution of actions and their corresponding categories.}}
    \label{supp:fig:sankey}
\end{figure*}

For the study, we curated 93 motion clips, organizing them by movement type and style into a structured dataset. We delineated movements into three categories: \emph{move\_around}, \emph{act\_in\_place}, and \emph{combined}; and styles into five categories: \emph{attack}, \emph{crawl}, \emph{jump}, \emph{dance}, and \emph{usual}. The clips were then classified into these eight categories, with a weighting system applied based on the inverse frequency of each category to enhance the representation of less common actions. For motions that spanned multiple categories, their weights were averaged based on their inverse frequency values. This approach aimed to ensure a balanced action distribution within the dataset, emphasizing the inclusion of rarer actions to avoid overrepresentation of any single action type. The categorization and its impact on the dataset distribution are illustrated in the diagram available in \cref{supp:fig:sankey}.

\section{Experiments}\label{supp:exp}

This supplementary section expands on the experimental analyses from \cref{sec:exp}, focusing on the text description. Beyond the quantitative metrics addressed in the main document, we explore the changes in reward function dynamics pre- and post-text refinement across various instructions. This includes a detailed comparison of CLIP similarity scores during training to critically evaluate the effectiveness and design of different reward functions.

\subsection{Text Enhancement}\label{supp:exp:text}

Utilizing the text enhancement strategy described in \cref{supp:textdata}, we have refined action descriptions from existing open-source datasets, reducing ambiguity and enhancing clarity and applicability. To gauge the impact of these refined descriptions on training efficacy, we track and compare the reward feedback during the training phases.

\begin{figure}[t!]
    \centering
    \begin{subfigure}[b]{0.5\linewidth}
        \includegraphics[width=\linewidth]{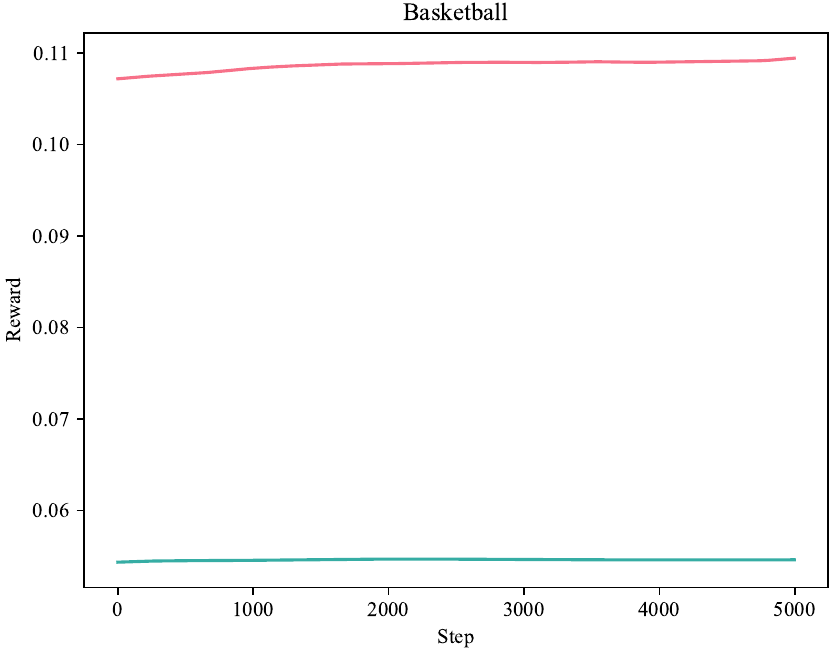}
    \end{subfigure}%
    \begin{subfigure}[b]{0.5\linewidth}
        \includegraphics[width=\linewidth]{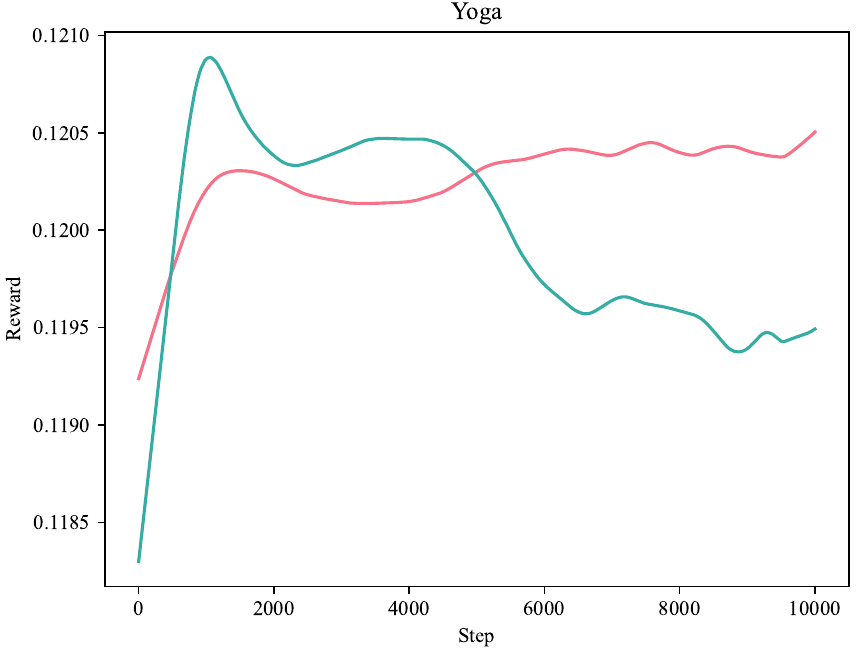}
    \end{subfigure}%
    \\%
    \begin{subfigure}[b]{0.5\linewidth}
        \includegraphics[width=\linewidth]{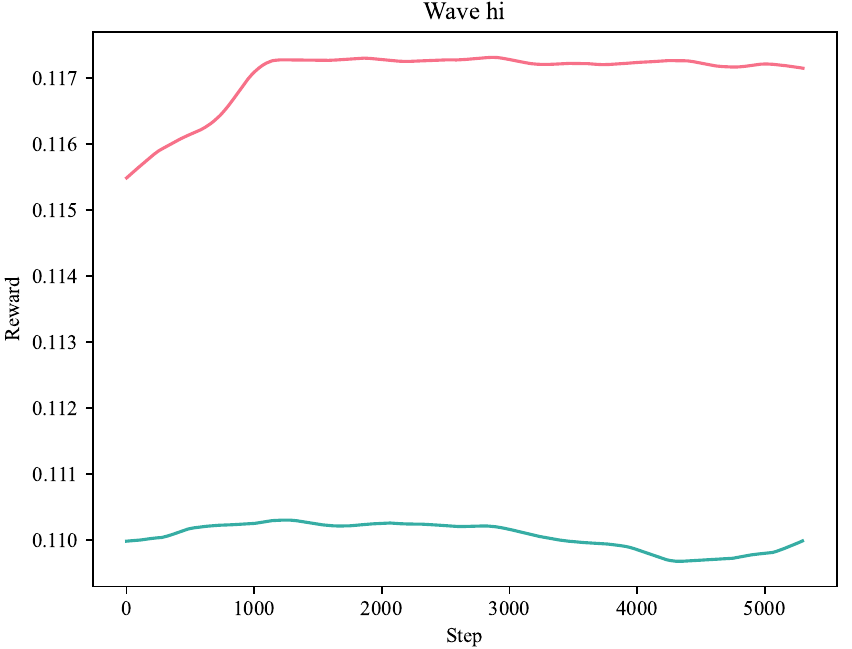}
    \end{subfigure}%
    \begin{subfigure}[b]{0.5\linewidth}
        \includegraphics[width=\linewidth]{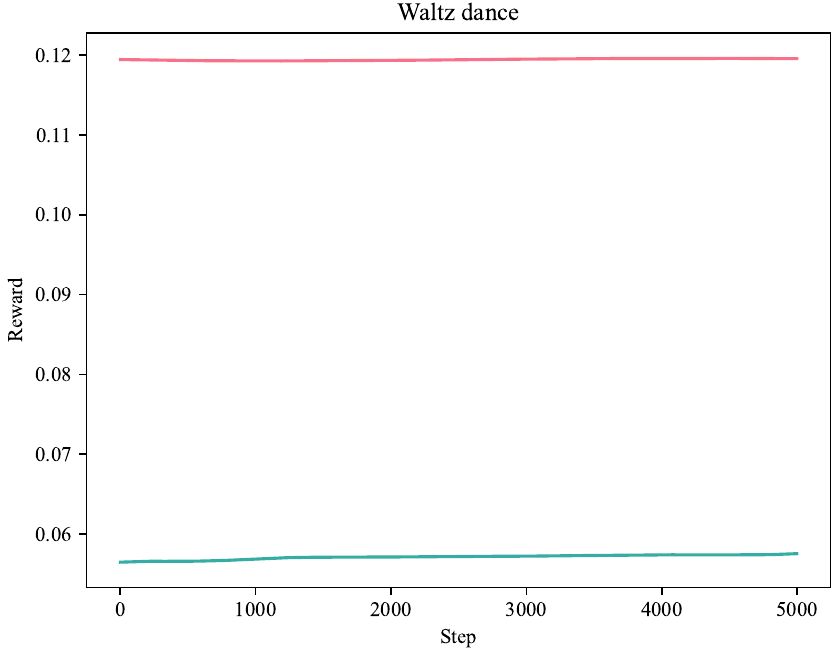}
    \end{subfigure}%
    \\%
    \begin{subfigure}[b]{0.4\linewidth}
        \includegraphics[width=\linewidth]{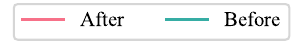}
    \end{subfigure}%
    \caption{\textbf{Rewards before and after text enhancement.} The red curve depicts reward trends following text enhancement, contrasting with the pre-enhancement trends shown by the green curve.}
    \label{supp:fig:ba}
\end{figure}

Selecting four instructions at random from our dataset for illustration, we compare reward trends before and after text enhancements---represented by green and red curves, respectively, in our graphs. This comparison reveals that refined instructions consistently yield superior reward trajectories from the start, showing a swift and steady ascent to a performance plateau. This indicates that text enhancement notably improves policy training efficiency and convergence speed. Specifically, for intricate actions like \emph{Yoga} (as shown in the top right figure of \cref{supp:fig:ba}), refined instructions result in a more stable and gradual reward increase, signifying improved training stability and model performance.

\begin{figure}[t!]
    \centering
    \begin{subfigure}[b]{0.5\linewidth}
        \includegraphics[width=\linewidth]{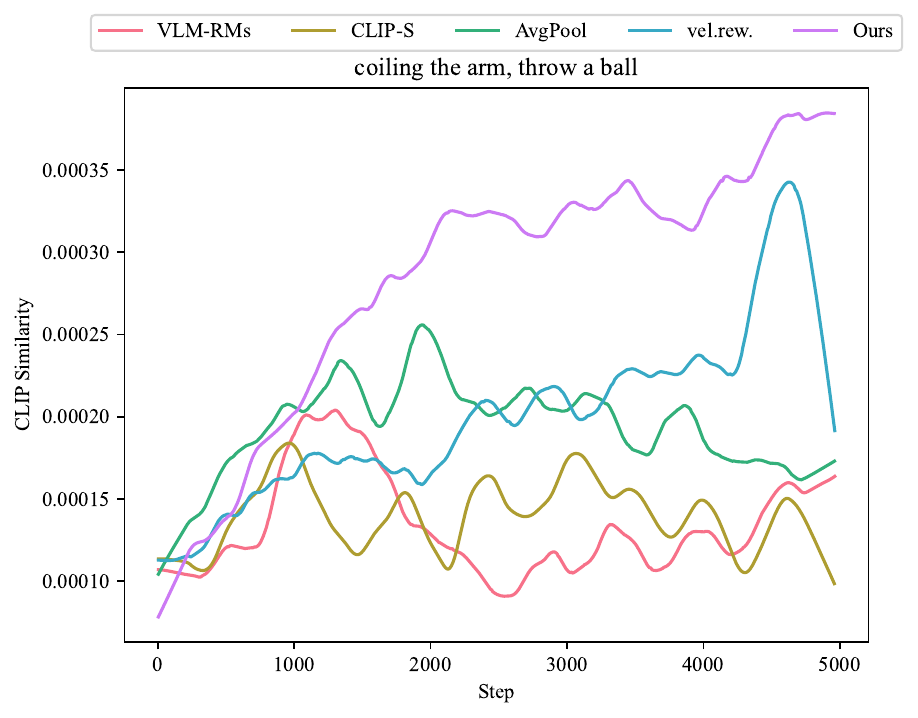}
    \end{subfigure}%
    \begin{subfigure}[b]{0.5\linewidth}
        \includegraphics[width=\linewidth]{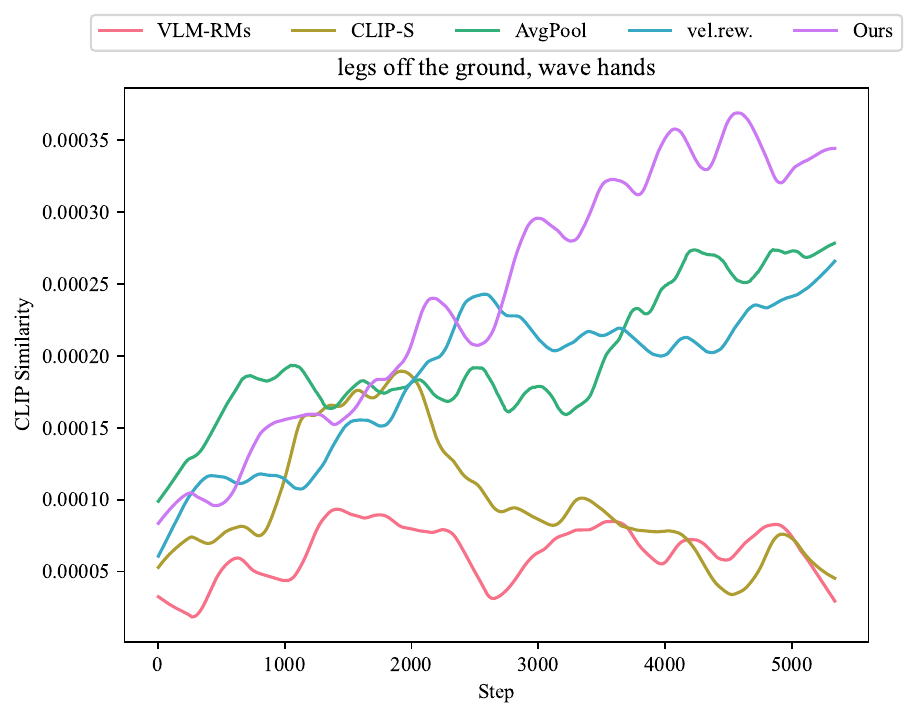}
    \end{subfigure}%
    \\%
    \begin{subfigure}[b]{0.5\linewidth}
        \includegraphics[width=\linewidth]{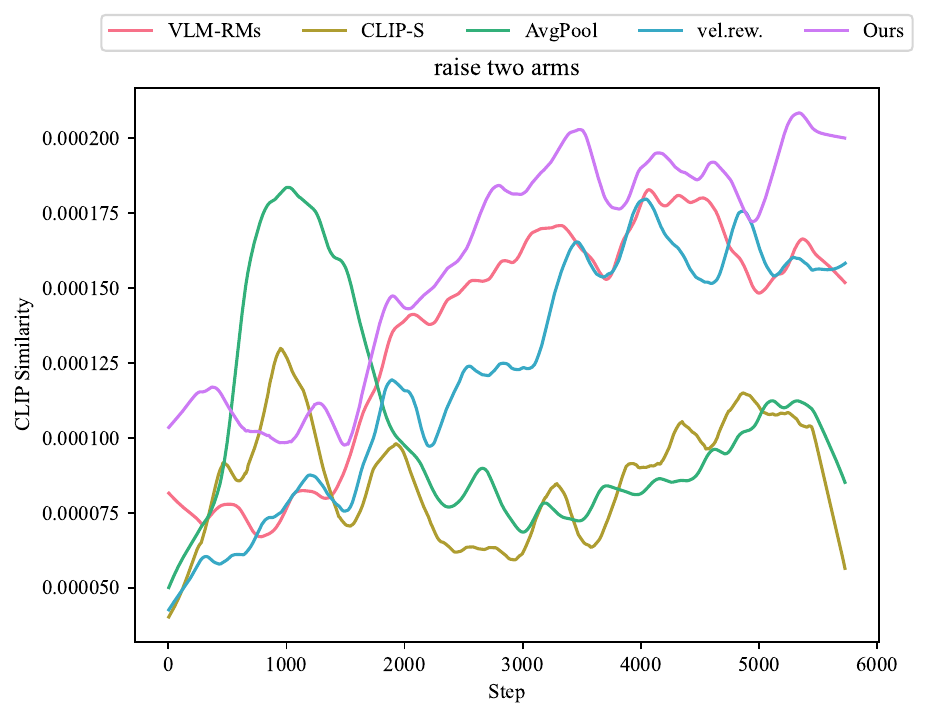}
    \end{subfigure}%
    \begin{subfigure}[b]{0.5\linewidth}
        \includegraphics[width=\linewidth]{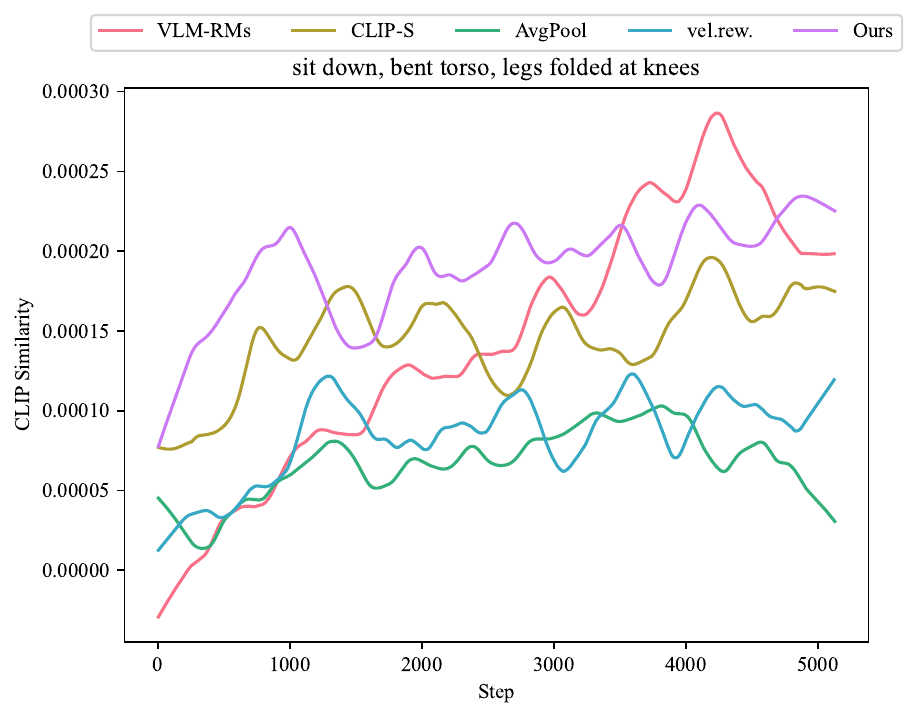}
    \end{subfigure}%
    \\%
    \begin{subfigure}[b]{0.9\linewidth}
        \includegraphics[width=\linewidth]{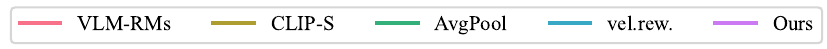}
    \end{subfigure}
    \caption{\textbf{The CLIP similarity calculated by different reward designs.}}
    \label{supp:fig:reward5}
\end{figure}

\subsection{Implementation Details}\label{supp:param}

\begin{figure*}[t!]
    \centering
    \begin{subfigure}[b]{\linewidth}
        \includegraphics[width=\linewidth]{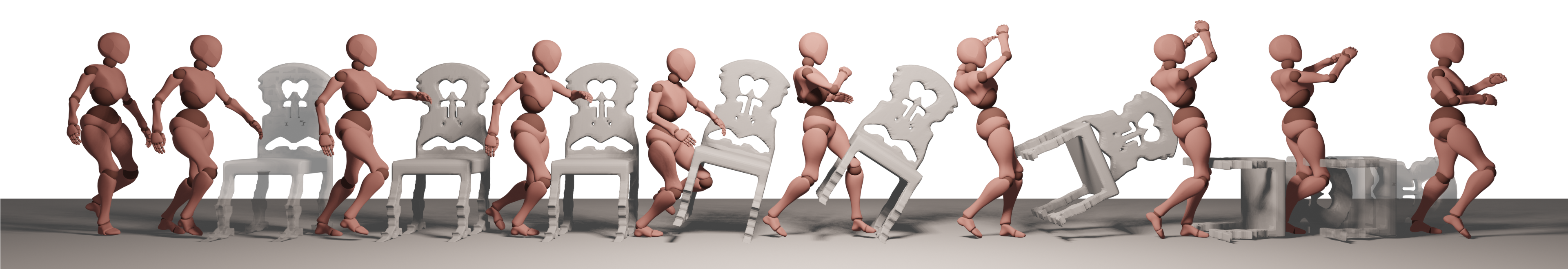}
        \caption{kick the white chair}
        \label{fig:chair1}
    \end{subfigure}%
    \\%
    \begin{subfigure}[b]{\linewidth}
        \includegraphics[width=\linewidth]{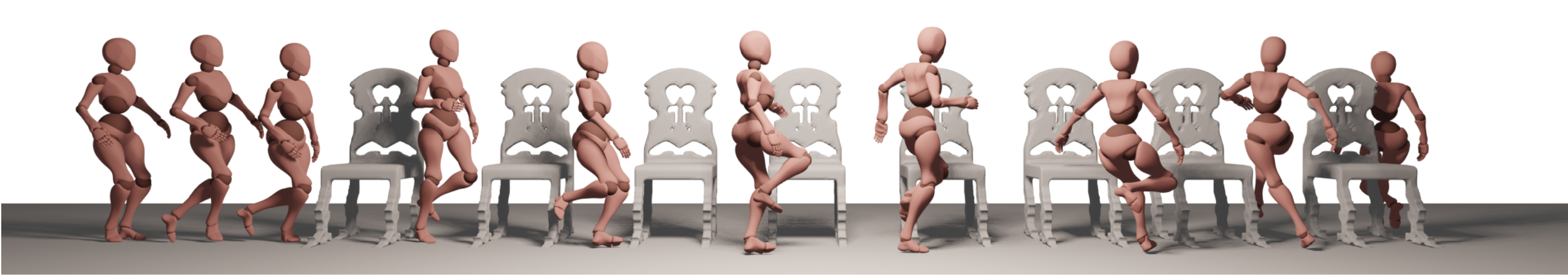}
        \caption{move around the white chair}
        \label{fig:chair2}
    \end{subfigure}%
    \\%
    \begin{subfigure}[b]{\linewidth}
        \includegraphics[width=\linewidth]{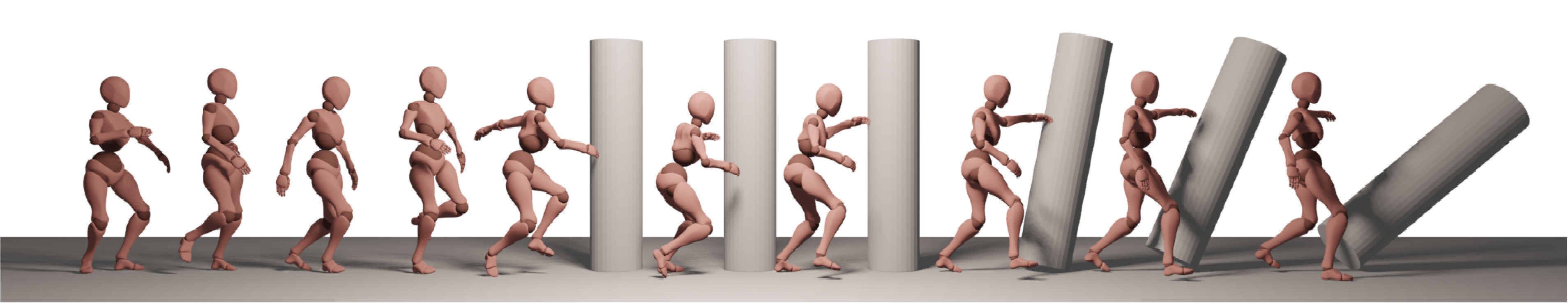}
        \caption{strike the pillar}
        \label{fig:pillar}
    \end{subfigure}%
    \label{supp:fig:interact}
    \caption{\textbf{Additional results of interaction motions.}}
\end{figure*}

\subsection{Reward Function Analysis}\label{supp:exp:reward}

To evaluate and compare various reward function designs, we use cosine similarity between image and text features as a uniform metric, accommodating the differing numerical scales inherent to each reward design. As depicted in \cref{supp:fig:reward5}, we represent five reward functions using distinct colors, with our method marked in purple.

Aligning with discussions in the main text (\cref{fig:high}), we examine four instructions from our user study for a detailed comparison. Our findings indicate that our method uniformly improves image-text alignment throughout training, achieving consistent convergence. While some methods exhibit comparable performance on select instructions, they generally show less consistency, with initial gains often receding over time. In contrast, our approach demonstrates robustness against the variabilities of open-vocabulary training, leading to stable and reliable performance improvements.

To assist readers in replicating our work, we have included a comprehensive breakdown of hyperparameter settings in \cref{supp:tab:low,supp:tab:high}. 

\begin{table}[ht!]
    \centering
    \small
    \caption{\textbf{Hyperparameters used for the training of low-level controller.}}
    \label{supp:tab:low}
    \begin{tabular}{ll}
        \toprule
        \textbf{Hyper-Parameters}         & \textbf{Values}       \\
        \midrule
        dim(Z) Latent Space Dimension         & 64       \\
        Encoder Align Loss Weight             & 1        \\
        Encoder Uniform Loss Weight           & 0.5      \\
        $w$ gp Gradient Penalty Weight        & 5        \\
        Encoder Regularization Coefficient    & 0.1      \\
        Samples Per Update Iteration          & 131072   \\
        Policy/Value Function Minibatch Size  & 16384    \\
        Discriminators/Encoder Minibatch Size & 4096     \\
        $\gamma$ Discount                     & 0.99     \\
        Learning Rate                         & $2 \times 10^{-5}$     \\
        GAE($\lambda$)                        & 0.95     \\
        TD($\lambda$)                         & 0.95     \\
        PPO Clip Threshold                    & 0.2      \\
        $T$ Episode Length                    & 300      \\
        \bottomrule
    \end{tabular}
\end{table}

\begin{table}[ht!]
    \centering
    \small
    \caption{\textbf{Hyperparameters used for the training of high-level controller.}}
    \label{supp:tab:high}
    \begin{tabular}{ll}
        \toprule
        \textbf{Hyper-Parameters}         & \textbf{Values}       \\
        \midrule
        $w$ gp Gradient Penalty Weight        & 5        \\
        Encoder Regularization Coefficient    & 0.1      \\
        Samples Per Update Iteration          & 131072   \\
        Policy/Value Function Minibatch Size  & 16384    \\
        Discriminators/Encoder Minibatch Size & 4096     \\
        $\gamma$ Discount                     & 0.99     \\
        Learning Rate                         & $2 \times 10^{-5}$     \\
        GAE($\lambda$)                        & 0.95     \\
        TD($\lambda$)                         & 0.95     \\
        PPO Clip Threshold                    & 0.2      \\
        $T$ Episode Length                    & 300      \\
        \bottomrule
    \end{tabular}
\end{table}

\subsection{Interaction Motions}\label{supp:interact}

Within the main text, we highlighted \model's proficiency in mastering tasks involving interactions with diverse objects, underscoring its capability to adapt across a spectrum of interaction scenarios. For experimental validation, we deliberately chose a range of objects, both rigid (\eg, pillars, balls) and articulated (\eg, doors, chairs), to demonstrate the method's versatility. The quantitative analyses of these object interactions, as detailed in \cref{supp:fig:interact}, affirm the flexibility of our approach. Our system is shown to adeptly navigate a variety of action requirements, as specified by different text descriptions, maintaining efficacy even when faced with repetitive initial conditions or identical objects.

\begin{figure}[htb!]
    \centering	
    \includegraphics[width=\linewidth]{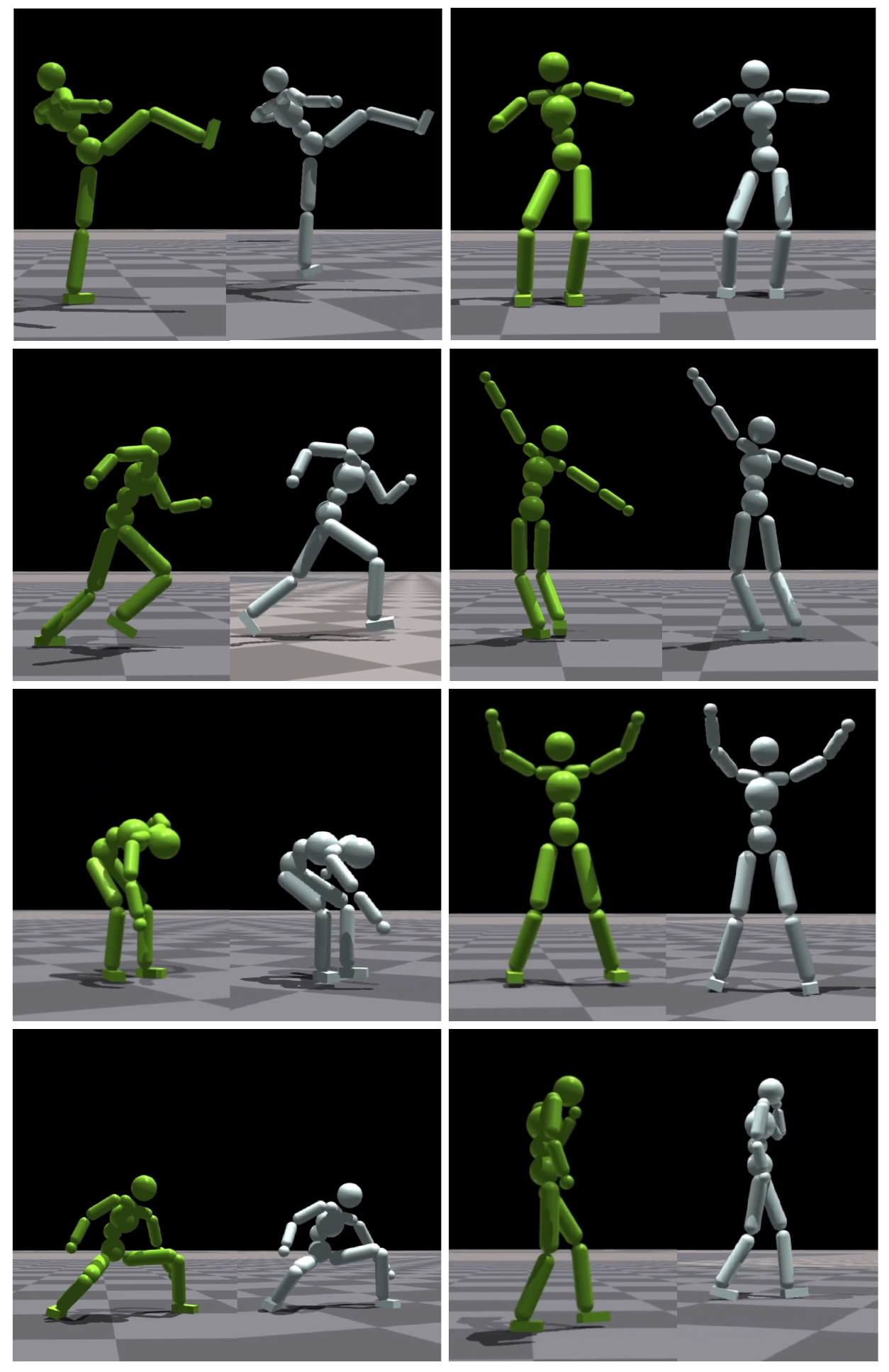}
    \caption{\textbf{Atomic actions from the trained low-level controller.} In each subfigure, the green agent shows the reference motion from the dataset, and the white agent shows our learned atomic action.}
    \label{supp:fig:low}
\end{figure}

\begin{figure*}[bth!]
    \centering	
    \includegraphics[width=\linewidth]{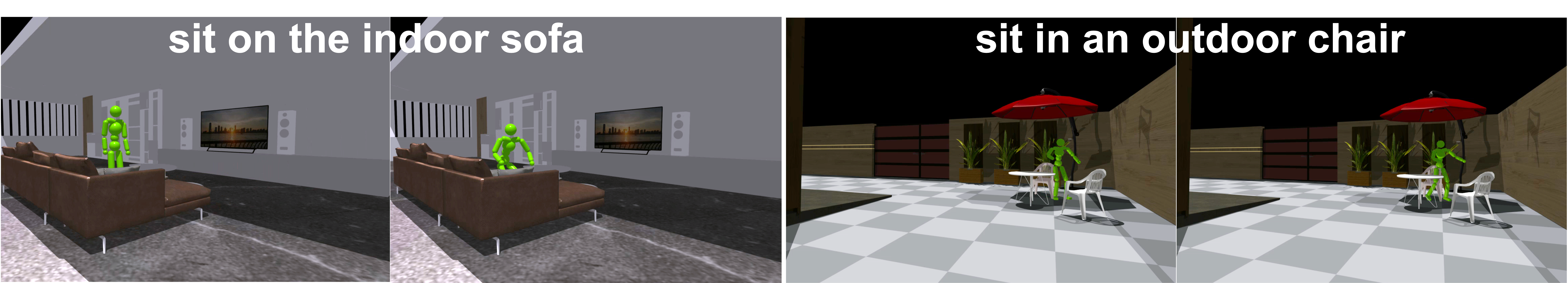}
    \caption{\textbf{Real-time scene interaction.} We employed both indoor and outdoor scenes within IsaacGYM. Throughout the training process, we conducted real-time rendering and obtained feedback on physical interactions.}
    \label{supp:fig:scene}
\end{figure*}

\clearpage

\begin{figure*}[t!]
    \centering
    \begin{subfigure}[b]{\linewidth}
        \includegraphics[width=\linewidth]{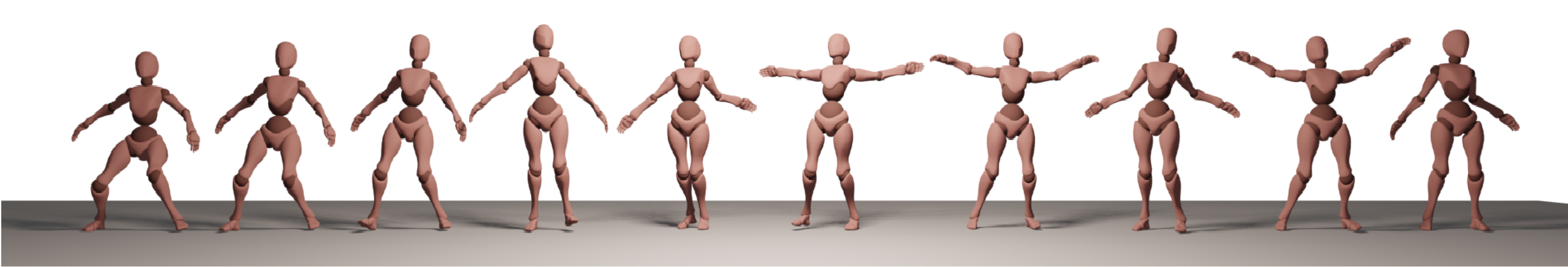}
        \caption{wave hands up and down}
        \label{supp:fig:wave}
    \end{subfigure}%
    \\%
    \begin{subfigure}[b]{\linewidth}
        \includegraphics[width=\linewidth]{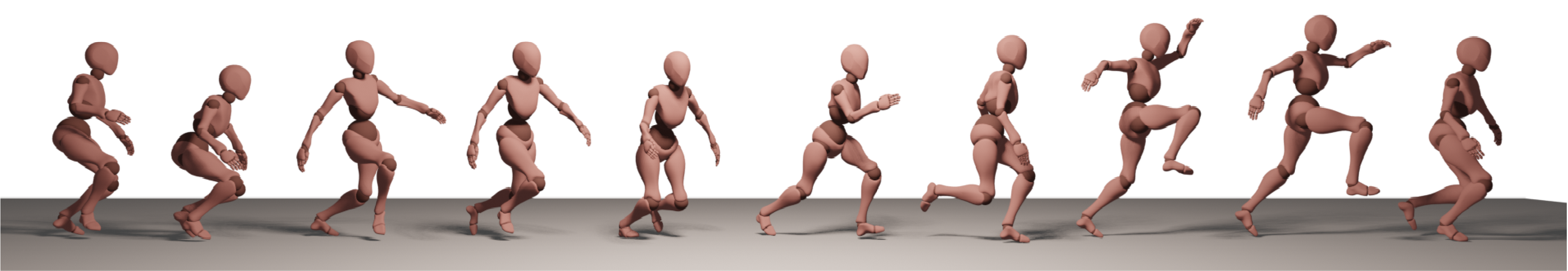}
        \caption{jump high}
        \label{supp:fig:jump}
    \end{subfigure}%
    \\%
    \begin{subfigure}[b]{\linewidth}
        \includegraphics[width=\linewidth]{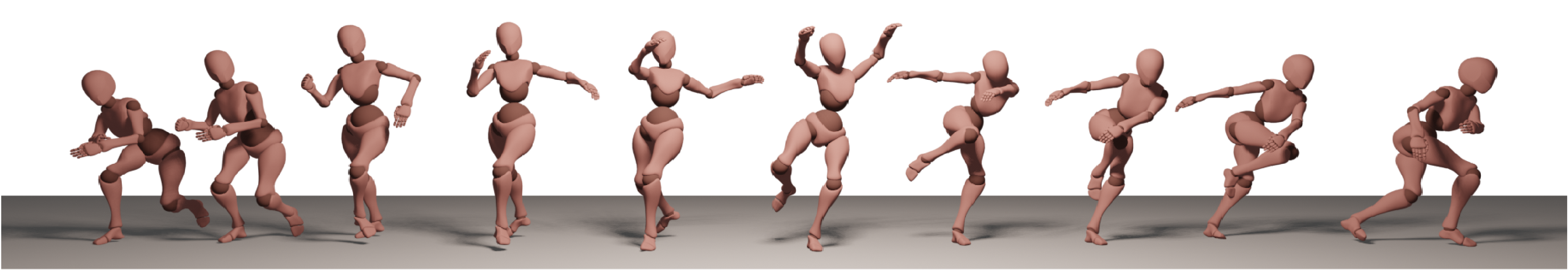}
        \caption{left leg forward, right leg retreats}
        \label{supp:fig:kick}
    \end{subfigure}%
    \\%
    \begin{subfigure}[b]{\linewidth}
        \includegraphics[width=\linewidth]{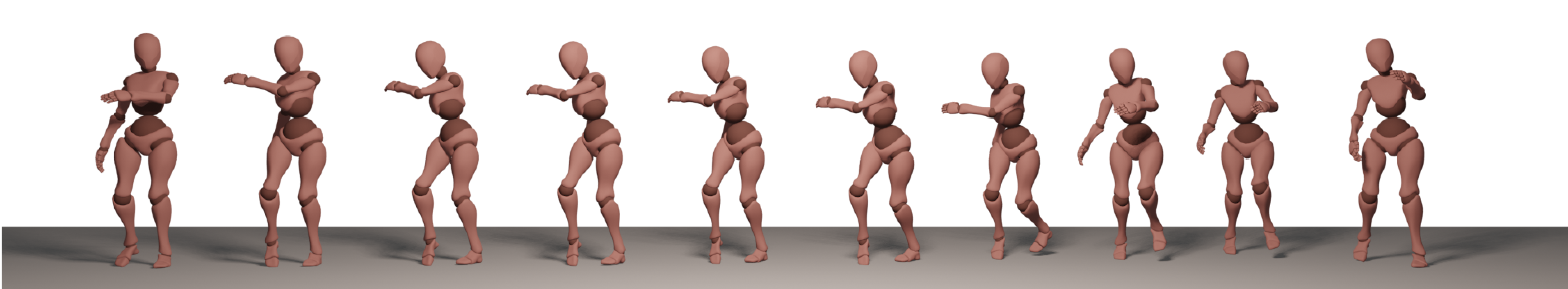}
        \caption{raise one arm, put the other hand down}
        \label{supp:fig:onearm}
    \end{subfigure}%
    \\%
    \begin{subfigure}[b]{\linewidth}
        \includegraphics[width=\linewidth]{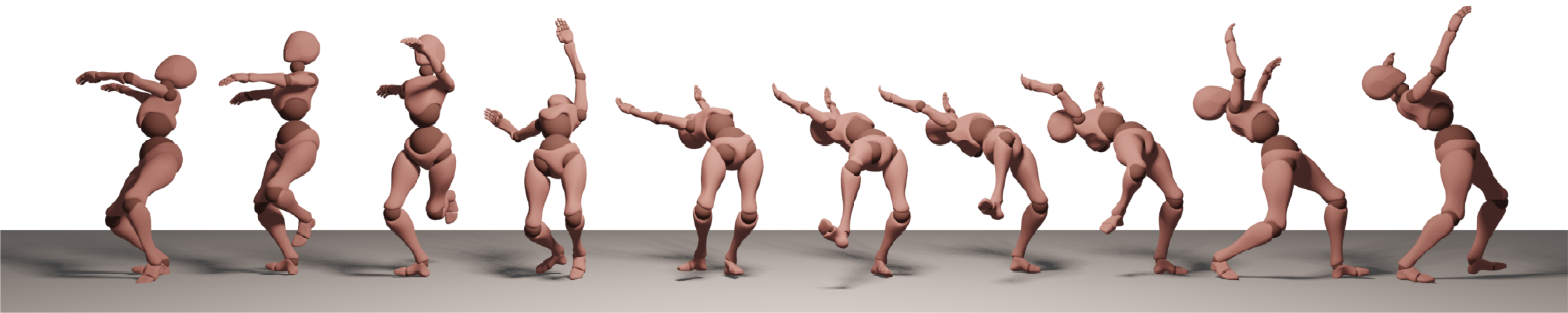}
        \caption{raise hands above head, bend body}
        \label{supp:fig:yoga}
    \end{subfigure}%
    \\%
    \begin{subfigure}[b]{\linewidth}
        \includegraphics[width=\linewidth]{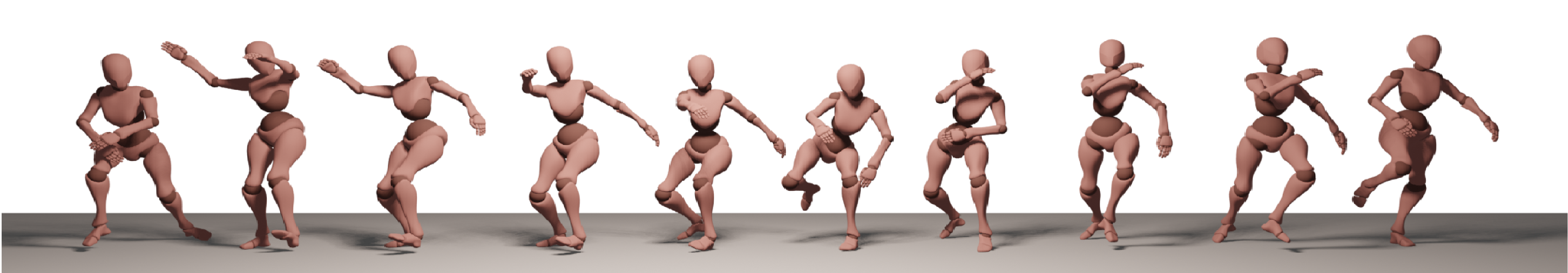}
        \caption{hit a tennis smash with arm}
        \label{supp:fig:tennis}
    \end{subfigure}%
    \\%
    \label{supp:fig:skill}
    \caption{\textbf{More results of open-vocabulary physical skills.}}
\end{figure*}

\clearpage

\end{document}